\documentclass[acmsmall]{acmart}
\AtBeginDocument{%
  \providecommand\BibTeX{{%
    \normalfont B\kern-0.5em{\scshape i\kern-0.25em b}\kern-0.8em\TeX}}}

\setcopyright{acmlicensed}
\copyrightyear{2018}
\acmYear{2018}
\acmDOI{XXXXXXX.XXXXXXX}

\acmJournal{JACM}
\acmVolume{37}
\acmNumber{4}
\acmArticle{111}
\acmMonth{8}

\usepackage{amsmath,amsfonts}
\usepackage{algorithmic}
\usepackage{algorithm}
\usepackage{array}
\usepackage[caption=false,font=normalsize,labelfont=sf,textfont=sf]{subfig}
\usepackage{textcomp}
\usepackage{stfloats}
\usepackage{url}
\usepackage{verbatim}
\usepackage{graphicx}
\usepackage{graphicx}
\usepackage{multirow}
\usepackage{booktabs}
\usepackage{amsfonts}
\usepackage{amsmath}
\usepackage{color,xcolor}
\usepackage{diagbox}
\usepackage{bbm}

\newcolumntype{L}[1]{>{\raggedright\let\newline\\\arraybackslash\hspace{0pt}}m{#1}}
\newcolumntype{C}[1]{>{\centering\let\newline\\\arraybackslash\hspace{0pt}}m{#1}}
\newcolumntype{R}[1]{>{\raggedleft\let\newline\\\arraybackslash\hspace{0pt}}m{#1}}
\newcommand{\tabincell}[2]{\begin{tabular}{@{}#1@{}}#2\end{tabular}}

\begin{document}

%%
%% The "title" command has an optional parameter,
%% allowing the author to define a "short title" to be used in page headers.
\title{DWE+: Dual-Way Matching Enhanced Framework for Multimodal Entity Linking}

%%
%% The "author" command and its associated commands are used to define
%% the authors and their affiliations.
%% Of note is the shared affiliation of the first two authors, and the
%% "authornote" and "authornotemark" commands
%% used to denote shared contribution to the research.
\author{Shezheng Song}
\email{ssz614@nudt.edu.cn}
\author{Shasha Li}
\authornote{Corresonding author}
\email{shashali@nudt.edu.cn}
\affiliation{%
  \institution{National University of Defense Technology}
  \city{Changsha}
  \state{Hunan}
  \country{China}
  \postcode{410073}
}

\author{Shan Zhao*}
\email{zhaoshan@hfut.edu.cn}
\affiliation{%
  \institution{Hefei University of Technology}
  \city{Hefei}
  \country{China}
}

\author{Xiaopeng Li}
\email{xiaopengli@nudt.edu.cn}
\author{Chengyu Wang}
\email{chengyu@nudt.edu.cn}
\affiliation{%
  \institution{National University of Defense Technology}
  \city{Changsha}
  \state{Hunan}
  \country{China}
}

\author{Jie Yu}
\email{yj@nudt.edu.cn}
\author{Jun Ma*}
\email{majun@nudt.edu.cn}
\author{Tianwei Yan}
\email{augusyan@hotmail.com}
\author{Bin Ji}
\email{jibin02@outlook.com}
\author{Xiaoguang Mao}
\email{xgmao@nudt.edu.cn}
\affiliation{%
  \institution{National University of Defense Technology}
  \city{Changsha}
  \country{China}}

% \author{Meng Wang}
% \affiliation{%
%   \institution{Hefei University of Technology}
%   \city{Hefei}
%   \country{China}
% }

%% By default, the full list of authors will be used in the page
%% headers. Often, this list is too long, and will overlap
%% other information printed in the page headers. This command allows
%% the author to define a more concise list
%% of authors' names for this purpose.
\renewcommand{\shortauthors}{Song et al.}

%%
%% The abstract is a short summary of the work to be presented in the
%% article.
\begin{abstract}
    Multimodal entity linking (MEL) aims to utilize multimodal information (usually textual and visual information) to link ambiguous mentions to unambiguous entities in knowledge base. 
    Current methods facing main issues:
    (1)treating the entire image as input may contain redundant information.
    (2)the insufficient utilization of entity-related information, such as attributes in images. 
    (3)semantic inconsistency between the entity in knowledge base and its representation. 
    %
    % Recently, we proposed a preliminary work of DWE to address the above issues: "A Dual-way Enhanced Framework from Text Matching Point of View for Multimodal Entity Linking" in AAAI 2024. It improves MEL by extracting the scene feature and leveraging the Wikipedia textual description as new entity representation. Despite the previous superior performance, we argue that scene feature may contain too much noise and static descriptions fail to maintain semantic consistency with the dynamic cognitive understanding of humans.
    % 
    To this end, we propose DWE+ for multimodal entity linking.
    DWE+ could capture finer semantics and dynamically maintain semantic consistency with entities. This is achieved by three aspects: 
    (a)we introduce a method for extracting fine-grained image features by partitioning the image into multiple local objects. Then, hierarchical contrastive learning is used to further align semantics between coarse-grained information(text and image) and fine-grained (mention and visual objects).
    (b)we explore ways to extract visual attributes from images to enhance fusion feature such as facial features and identity.
    (c)we leverage Wikipedia and ChatGPT to capture the entity representation, achieving semantic enrichment from both static and dynamic perspectives, which better reflects the real-world entity semantics.
    % % (1) 提取更相关的identity信息，减少噪声 （2）层次对比loss让模型可以从整体和细致两层语义来进行学习 （3）动态增强的method保持语义一致性
    % Firstly, we introduce a method for extracting fine-grained image features by partitioning the image into multiple local regions. Additionally, we utilize pre-trained models or image recognition tools to extract attributes with finer granularity and stronger relevance from images, such as facial features and identity.
    % Secondly, we improve the methods for enhancing entity representation: static enhancement and dynamic enhancement. These methods effectively reduce the disparity between representation and entities.
    Experiments on Wikimel, Richpedia, and Wikidiverse datasets demonstrate the effectiveness of DWE+ in improving MEL performance. Specifically, we optimize these datasets and achieve state-of-the-art performance on the enhanced datasets. The code and enhanced datasets are released on \url{https://github.com/season1blue/DWET}.
\end{abstract}

%%
%% The code below is generated by the tool at http://dl.acm.org/ccs.cfm.
%% Please copy and paste the code instead of the example below.
%%


%% Keywords. The author(s) should pick words that accurately describe
%% the work being presented. Separate the keywords with commas.

\received{20 February 2007}
\received[revised]{12 March 2009}
\received[accepted]{5 June 2009}

\maketitle

\section{Introduction}

\begin{figure}[pbt]
        \centering
        \includegraphics[width=.7\linewidth]{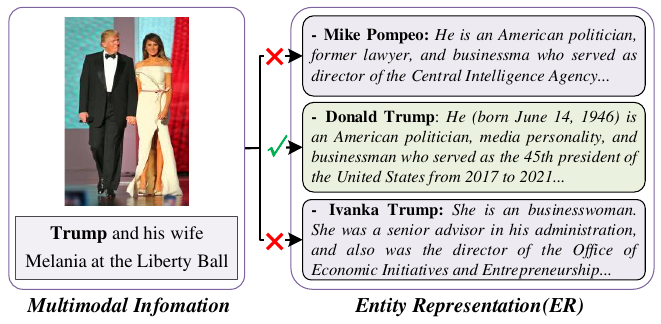}
        \caption{Example of entity linking for the mention \textit{Trump}.}
        \label{fig:intro1}
    \end{figure}

    Entity linking (EL) \cite{el1, MEGCF} has attracted increasing attention in the natural language processing (NLP) domain, which aims at linking ambiguous mentions to the referent unambiguous entities in a given knowledge base (KB) \cite{ kb1}. It is crucial in many information retrieval applications, such as information extraction \cite{ie3}, question answering \cite{intro_qa1, dialog}, recommendation\cite{DynamicMultimodal} and Web query \cite{intro_sq3, Search}. However, the ambiguity in mentions poses great challenges to this task. Hence, multimodal entity linking (MEL) has been introduced recently \cite{wikiperson, mel_method2}, where multimodal information (e.g., visual information) is used to disambiguate the mention.
    Figure \ref{fig:intro1} shows an example of MEL, wherein the mention \textit{Trump} is associated with the text \textit{``Trump and his wife Melania at the Liberty Ball"}. It is challenging to distinguish the entity because the short text is ambiguous, and several entities are related to \textit{Trump}. Thus, image is required for further disambiguation. 
    Dozens of MEL studies have been conducted in the past few years and achieved promising performances, where delicate interaction and fusion mechanisms are designed for encoding the image and text features \cite{baseline_mel,baseline_dataset, hybrid}. Nonetheless, existing approaches still struggle to fully exploit the potential of the feature sources from two distinct information perspectives, which may hinder further multimodal task development.

    % 图像利用度不足。
    % 现有方法对于图像利用存在两个缺点，i.现有方法通常会将整个图像作为信息输入，一些考虑精细化图像信息利用的方式如VIT，会将图像进行均等分割16*16的小块进行输入。但是对于实体链接任务中的图像信息来说，重要信息不是均匀分布的，而是object-oriented，因此VIT那种平均分割的方式对于MEL的表现不佳。ii.MEL任务中的图像中最重要的信息是关于实体的属性信息。以人物实体为例，图中最重要的信息是人物的年龄、肤色等能够帮助确定实体独特性的信息，但是现有方法对于这部分信息的提取能力和利用能力不足，现有方法通常依赖于在通用化的图像信息上预训练的编码器对图像进行处理，这种编码方式弱于对实体独特性信息的捕捉。
    
    \textbf{\textit{Insufficient Utilization of Images}}
    The existing methods have drawbacks in utilizing images: 
    \textbf{(a)} Treating the entire image as input may contain redundant information.
    In MEL task, one of the challenges lies in the fact that the visual modality often contains less information~\cite{kim2020modality} and more redundant features when compared to the text modality.  Indeed, not all visual sources play positive roles. As revealed by \citet{vempala2019categorizing}, as high as 33.8\% of visual information serves no context or even noise in task. Current methods~\cite{resnet, baseline_mel, wikidiverse, MMEL} typically utilize the entire image as input information. However, this approach tends to input redundant information as well. 
    Some approaches, such as ViT~\cite{ViT, touvron2021training}, which consider fine-grained image information, divide the image into equally sized 16*16 patches for input. However, for image information in entity linking tasks, important information is not uniformly distributed but object-oriented. The parts of the image related to entity objects are the ones that should be emphasized. As shown in Figure \ref{fig:intro1},  only the left person is relevant while the right person is noise, causing redundant visual elements and low information density for the entity. 
    Therefore, directly using the coarse-grained method such as ViT could lead to information mismatches and is not optimal for MEL performance. We argue that a fine-grained and object-oriented visual feature is needed.
    \textbf{(b)} The insufficient utilization of entity-related information, such as attributes in images\cite{anpextractor, liu2020exploring}.
    Due to the definition and requirements of the MEL task, the most important information in images is about entity attributes. Taking a person entity as an example, the most crucial information in the image includes the person's age, skin color, and other attributes that help determine the entity's uniqueness~\cite{deepface}. However, current methods cannot extract and utilize this information effectively. They usually rely on encoders~\cite{clip_model, resnet, ViT} pretrained on generalized image data to process images, which are weaker in capturing entity-specific information.

    % 现存ER表示方法与知识库中实体语义的不一致
    \textbf{\textit{Inconsistency between entity and representation:}}
    % 人类在构建知识库的时候是基于自身对于实体的认知来构建实体的表示，同时在日常的多模态表达中，也是基于人类对于实体的语义认知而表达。所以MEL任务的核心点就是要利用这种语义一致性，来捕捉多模态信息和知识库实体中的共同点。但是当实体表示与实体语义存在分歧时，这种一致性就无法被利用，导致模型的学习性能受到影响。
    Humans construct representations of entities in the knowledge base based on their cognitive understanding of those entities. Similarly, people's expression of mentions is based on their semantic understanding of entities. Therefore, the core of the MEL task is to utilize this semantic consistency to capture commonalities between entities in the knowledge base and mention. 
    As shown in Figure \ref{fig:framework}, entity representation(ER) serves as the bridge linking mentions and entities. Mentions are first linked to ER which represents the corresponding entity, thus enabling the matching between mentions and entities. Hence, consistency among ER, entity semantics, and mention-related information is essential. Otherwise, even if the linkage between mention and ER is accomplished, the linkage between mention and entity remains incorrect. However, existing methods fail to achieve consistency between ER and entity semantics, which leads to a deviation in the model's learning direction.
    For instance, \citet{baseline_mel} and  \citet{baseline_dataset}  concatenate the main properties of an entity into a piece of text, which is regarded as a representation of the entity in KG. \citet{wikiperson}manually collect short text to represent entity, e.g., ``Bahador Abb: Iranian footballer". However, the set of properties or short text cannot represent entities. Different people may have the same properties. For example, the entity \textit{Donald Trump} and \textit{Mike Pompeo} both have the properties of \textit{``Sex: male. Religion: Presbyterianism. Occupation: businessperson, politician. Languages: English. work location: Washington, D.C."}
    With so similar ER, it is a great challenge to link entities even with an ideal multimodal method. Thus, a more distinctive and representative ER is required for further disambiguation. Fortunately,  Wikipedia description and ChatGPT both offer promising solutions, which have been shown to enrich the semantics of raw data, especially in low-resource NLP tasks \cite{chen2022few}.
    
                               %%%%%%%%%%%%% 如何解决问题 %%%%%%%%%%%%%%%%%%
                               %%%%%%%%%%%%% 如何解决问题 %%%%%%%%%%%%%%%%%%
                               
    To tackle the above issues, we propose a novel framework to improve  MEL.
    Inspired by the neural text matching task, we view each mention as a query and attempt to learn the mapping from each mention to the relevant entity. To achieve this, we design a dual-way enhanced \textbf{(DWE+)} framework: enhancing query with fine-grained image and visual attributes, and enriching the semantics of entity by static or dynamic method:

    \textbf{ (1)}
    %我们采取了两种方法来增强图像的利用度。
    % 首先，我们对图像进行了精细化的利用。根据目标检测的结果，我们将整个图像划分为多个局部区域，并以Object为导向地提取图像信息。图像可以被分解为多个局部区域，每个实体提及只描述一个局部区域。我们使用对象检测器识别图像中的所有可能对象，并通过CLIP的图像编码器提取对象的视觉特征，进行相应的利用。
    % 另一方面，我们使用预训练模型或图像识别工具对图像进行预处理，以提取其中的相关信息，如面部信息、身份信息等。具体地，我们使用面部特征检测工具，如Deepface，对分割后的目标图像进行预处理，以提取面部外观，例如性别、种族、年龄等，这些与图像中面部特征的属性相关。同时，我们使用预先训练的ViT模型对图像进行预处理提取身份信息，该模型是在人脸识别数据集MS-Celeb1M上进行预训练的，以分析图像中的人物可能与哪些名人相似，为后续实体识别提供辅助信息。我们通过多种方式显式地提取了有效属性帮助实体链接。然而，考虑到提取的准确性，模型需要在加入辅助信息时进行适当的考虑。 
    The DWE+ employs a pretrained visual encoder and visual tools to obtain the image representation and fine-grained visual attributes to enhance the utilization of images.
    \textbf{(a)}
    we introduce a method for fine-grained image features by partitioning the image into multiple local objects and extracting object-oriented image information. The image can be decomposed into multiple local regions, with each entity describing only one local region. Object detector is leveraged to identify all possible objects in the image and extract the visual features of objects through CLIP. 
    Then, hierarchical contrastive learning is used to further align semantics, including coarse-grained and fine-grained contrastive learning. Coarse-grained contrastive learning aims to guide the alignment of overall semantics (text and image), while fine-grained contrastive learning aims to guide the alignment of target-relevant information (mention and visual objects).
    \textbf{(b)}
    we explore ways to extract visual attributes from images to enhance fusion feature such as facial features and identity.
    % We preprocess the images using pre-trained models or image recognition tools to extract relevant information such as facial attributes and identity information. 
    Specifically, we use facial detection tools such as Deepface to preprocess the segmented target image and extract facial attributes such as \textit{gender, race, and age}, which are relevant to facial appearance in the image. Additionally, we employ a pre-trained ViT model for preprocessing images to extract identity information. This ViT is pre-trained on the MS-Celeb1M face recognition dataset to analyze which celebrities the people in the images may resemble. In summary, We explicitly extract effective attributes through various methods to assist entity linking. 
    % However, considering the accuracy of extraction, it needs to be considered appropriately when incorporating auxiliary information.
    
    % \textbf{(1)} For the enhancing query (e.g., mention), DWE employs a pretrained visual encoder and visual tools to obtain the image representation and fine-grained visual attributes, then purify it into the learnable visual characteristics. The visual characteristics and text are regarded as two types of refined information, which are separately employed to enhance the mention.  Considering the semantic gap between vision and text in the language model, we design three enhanced units based on the cross-modal enhancer to bridge the semantic gap between (i) facial feature and object visual feature (ii) mention and vision-enhanced feature and (iii)mention and text-enhanced feature. By introducing the cross-modal alignment, our model makes use of the semantics of text and objective image information.

    \textbf{ (2)} 
    % 我们探究了两种增强实体与representation一致性的方法，为了更好地使得representation贴近实体的真实语义，我们提出了（1）静态增强方法。基于wiki百科中实体页面中的静态文本来完成实体表示的扩充，其特点是准确且清晰（2）动态增强方法。依赖于飞速发展的大模型，我们使用多轮询问的方式从大模型中获取实体的对应表示。由于大模型如GPT4是动态和实时更新的，所以其实体表示也是动态，更贴合实际世界中实体信息随时间更新的情况。
    We explore two methods to enhance the consistency between entities and representations. To better align entity representation(ER) with the true semantics of entities, we propose the methods: 
    \textbf{(a)} Static enhancement method. This method relies on the static text from entity pages in Wikipedia to augment ER, ensuring enhanced accuracy and clarity. 
    \textbf{(b)} Dynamic enhancement method. Leveraging the rapid development of large-scale models, we employ a multi-round questioning approach to obtain corresponding representations of entities from large language models(LLM). Since LLMs like GPT-4 are dynamic and continuously updated, the entity information in their knowledge is also dynamic, better reflecting the real-world scenario where entity information evolves.
    Both methods reduce the disparity between textual representation and the entities in the knowledge graph, resulting in a more comprehensive textual representation. To sum up, we enhance ER for  17391, 17804 and 57007 entities in the Wikimel, Richpedia and Wikidiverse datasets, respectively.

    We note that a shorter conference version of this article\cite{dwe} is accepted for AAAI 2024.
    % 我们原本的方法主要是针对于双向增强的方法，初步探索了增强entity representation的方式以及使用多模态信息与mention信息交互以增强特征。然而，我们argue that dual way enhancing method is beneficial for multimodal entity linking. 为了进一步完成双向增强，一部分，我们细致分析了现有entity representation的特点，并提出了两种增强方式及增强后的数据集：静态增强和增强。另一部分，我们进一步针对图像中的人物显式的提出更多属性辅助完成实体链接，如人物身份信息，利用层次化对比学习进一步完成整体语义和目标对象语义的对齐，分别从的全局性和局部性来优化模型的学习。本文同样提供了在增强前后的数据集上的实验及分析。
    Our initial conference paper is only an initial exploration of enhancing entity representation and leveraging multimodal information to enhance features. However, we argue that previous scene feature may contain too much noise, and static descriptions fail to maintain semantic consistency with the dynamic cognitive understanding of humans.
    To this end, we propose DWE+, an improved method of DWE.
    To further accomplish bidirectional enhancement, DWE+ meticulously analyzes the characteristics of existing entity representation and proposes two methods along with enhanced datasets: static enhancement and dynamic enhancement. Additionally, DWE+ explicitly extracts more attributes in images to assist in entity linking, such as personal identity information. Furthermore, DWE+ utilizes hierarchical contrastive learning to further align overall and target object semantics, optimizing model learning from global and local perspectives. This paper also provides experiments and analysis on datasets before and after enhancement.

    % 实验证明我们的方法在传统数据集Wikimel、Richpedia和Wikidiverse上优于大部分模型，并且在我们所提出的两种增强方法上优于所有的基线模型，实现了SOTA
    The extensive experiments demonstrate that our method outperforms most models on the original datasets Wikimel, Richpedia, and Wikidiverse. Furthermore, it surpasses all baseline models on both of our proposed enhanced datasets(static and dynamic), achieving state-of-the-art(SOTA) results.
    Our contributions are as follows:
    
    % \item We explore Multimodal Entity Linking in a neural matching formulation. We design a dual-way enhanced framework: query is enhanced by refined multimodal information and we utilize Wikipedia to enrich the semantics of entity representation.
    % \item We innovatively introduce fine-grained image attributes (such as facial characteristics and scene feature) which are leveraged to enhance and refine the visual features and take cross-modal alignment to bridge the semantic gap between textual and visual features. 
    % \item Experimental results on three benchmark datasets Richpedia, Wikimel and Wikidiverse demonstrate the superiority of our proposed model over the SOTA method.
    
    % 对于多模态实体链接任务，我们从实体表示和现实世界实体语义的一致性角度出发，优化实体表示的方式，探索了两种更佳的实体表示方式，使得实体表示更加能够代表知识库中的实体。
    % 我们从特征层次使用精炼后的图像特征对模型进行优化， We innovatively introduce fine-grained image attributes (such as facial characteristics and scene feature) which are leveraged to enhance and refine the visual features and take cross-modal alignment to bridge the semantic gap between textual and visual features. 
    % 我们使用优化后两种实体表示方式对Wikimel, richpedia and wikidiverse数据集进行了优化：使用固态方式进行增强的wiki-s,rich-s,diverse-s， 以及使用动态方式进行增强的wiki-d, rich-d, diverse-d.我们将会公开优化后的数据集。
    % 我们对我们的方法在三个数据集上进行了测评，验证了我们方法的有效性。更进一步的，我们将DWE方法在优化后的六个数据集上都进行了实验，demonstrate the superiority of our proposed model over the SOTA method.
    \begin{itemize}
        \item To mitigate the issue caused by redundant information and insufficient utilization of image, we propose DWE+. we revisit the multimodal entity linking as text matching by exploring ways to extract visual attributes from images to enhance fusion feature and build dynamic entity representation.
        \item Hierarchical contrastive learning is used to facilitate cross-modal alignment including coarse-grained (text and image) and fine-grained (mention and visual objects) contrastive learning, leading the model to learn from both global semantics and local semantics perspectives.
        % \item From the perspective of consistency between entity representations and real-world entity semantics, we explore two improved ways of optimizing entity representations to better signify the entity's semantics in the knowledge base.
        \item We explore and apply the two optimized entity representation approaches to enhance the Richpedia, Wikimel, and Wikidiverse datasets. Specifically, we employ a static enhancement method to get Rich-S, Wiki-S, and Diverse-S datasets; and a dynamic enhancement method to get Rich-D, Wiki-D, and Diverse-D datasets. The optimized datasets are released to the public.
        \item We evaluate DWE+ on original datasets (i.e. Richpedia, Wikimel, and Wikidiverse), confirming its effectiveness. Furthermore, we conducted quantitative experiments on the six optimized datasets, demonstrating that DWE+ could achieve state-of-the-art(SOTA) performance.
    \end{itemize}

\section{Related Work}
    MEL has been widely studied in recent years, which leverages the associated multimodal information to better link mention to entity in KG. Previous studies mainly focus on the following two aspects:

    % 讲数据集的缺乏和不足
    \textbf{Ambiguous Entity Representation}
    % Dozens of studies have proposed various MEL datasets for research. \citeauthor{tweet} \cite{tweet} propose a framework for automatically building the MEL dataset from Twitter\cite{twitter}. \citeauthor{weibo} \cite{weibo} study on a Chinese MEL dataset collected from the Chinese social media platform, namely, Weibo, which mainly focuses on the person entities. \citeauthor{movie} \cite{movie} release a MEL dataset collected from movie reviews, mainly focusing on film and film characters. \citeauthor{baseline_dataset} \cite{baseline_dataset} propose three MEL datasets, built from Weibo, Wikipedia, and Richpedia information and use CNDBpedia \cite{CNDBpedia}, Wikidata \cite{Wikidata2}, and Richpedia \cite{Richpedia} as the corresponding KG. However, the entity representation of these datasets is ambiguous, making it hard to distinguish the MEL target (entity). This paper differs from the above works in that we use the textual Wikipedia description as entity representation to guide the learning of multimodal feature, which are proven to be more efficient in our experiments.
    It is still controversy over how to represent the entity in KG, so there are different methods of entity representation. \citeauthor{tweet} \cite{tweet} propose a framework for automatically building the MEL dataset from Twitter\cite{twitter}, taking profile on Twitter as the user entity representation. However, the profile is edited by user and can not fully represent the user entity. \citeauthor{baseline_dataset} \cite{baseline_dataset} propose three MEL datasets, built from Weibo, Wikipedia, and Richpedia information, collecting the property of entity as representation, such as ``Donald Trump: male, 1946, politician" and using Wikidata \cite{Wikidata2} and Richpedia \cite{Richpedia} as the corresponding KGs. Similarly, Wikiperson\cite{wikiperson} dataset takes a manually organized short text as entity representation such as ``Bahador Abdi: Iranian footballer".
    However, the entity representation of the above datasets is ambiguous, making it hard to distinguish the MEL target (entity) and affecting the plausibility of performance evaluation.
    Our work differs from the above papers in that we use the textual Wikipedia description as entity representation to guide the learning of multimodal feature, which are proven to be more efficient.
    
    % 讲MEL的应用前景和范围 MEL的做法
    \textbf{Multimodal Information Refinement}
    Recently, dozens of works have focused on utilizing multimodal information for disambiguation. Wang et al. \cite{baseline_mel} propose to use gated multimodal feature fusion and contrastive learning for MEL. \citeauthor{wikidiverse} \cite{wikidiverse} use ResNet\cite{resnet} as a visual encoder and BERT\cite{BERT} as a textual encoder to extract the joint feature for MEL. However, these studies~\cite{zhao2023mcl, zhao2021enhancing, zhao2021dynamic} ignore the noise of the image, i.e., some objects in the image are irrelevant to the mention. \citeauthor{hybrid} \cite{hybrid} notice the visual and textual noise and take image segmentation and attention mechanism as refinement method, but ignore the global relationship between text and image. \citeauthor{ita} \cite{ita} also realize the noise, but they pay too much attention to extracting text such as OCR text or caption from raw images. To conclude, these work~\cite{ma2023using, ma2022joint} can not fully use multimodal information and filter the irrelevant part (i.e., noise) of raw image effectively. 
    % Differing from these works, DWE utilizes multimodal information by hierarchical contrastive learning and prompt learning. Thus DWE can effectively learn better joint representation.

\section{Factualized Entity Representation}
    \begin{figure}[b]
        \centering
        \includegraphics[width=.85\linewidth]{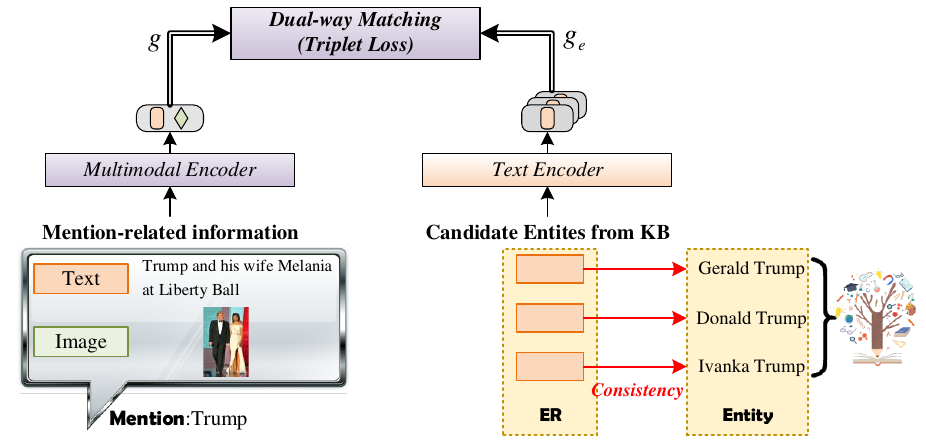}
        \caption{Dual-way Matching between mention and entity.}
        \label{fig:framework}
    \end{figure}
% 现在的ER是啥问题，如何构建一个ER
    Entity linking significantly influences and facilitates a profound understanding and information. 
    % 如下所示为实体链接任务的框架图。实体链接任务是一个双向匹配问题，旨在从数据库中找到与提及相关的实体。值得注意的是，需要确保Entity Representation（ER）与实体之间的语义一致性。右侧的候选实体是以Entity Representation作为输入的。如果模型能够成功将左侧的多模态信息与ER进行链接，则视为完成对实体的链接。然而，如果ER和实体之间存在不一致的情况，则可能导致链接任务失败和模型学习方向的偏差。因此，一个符合实际、符合实体真实语义的Representation是必要的。
    The entity linking framework diagram is illustrated in Figure \ref{fig:framework}. Entity linking is a bidirectional matching problem aimed at identifying entities relevant to mentions from a knowledge base. It is essential to ensure semantic consistency between entity representation (ER) and entities. The candidate entities on the right are inputted based on ER. Successful linking of multimodal information on the left with ER is considered as completing the entity linking task. However, inconsistencies between ER and entities can lead to task failure and model learning biases. Therefore, a representation that aligns with reality and reflects the true semantics of entities is necessary.

\begin{table}[tbp]
  \small
  \centering
  \caption{Examples of entity representations for the entities \textbf{Donald Trump} and \textbf{Barack Obama} in different datasets. \textbf{Wiki-S} and \textbf{Wiki-D} is the proposed dataset with static and dynamic enhancement, respectively.}
    \begin{tabular}{p{6.21em}|p{15em}|p{15em}}
    \toprule
    \multicolumn{1}{r|}{} & \multicolumn{1}{c|}{Donald Trump} & \multicolumn{1}{c}{Barack Obama} \\
    \midrule
    Wikimel\cite{baseline_mel} & Sex: male. Birth: 1946, Jamaica Hospital. Religion: Presbyterianism. Occupation: business magnate, entrepreneur, game show host, politician. Spouse: Melania Trump. Languages: American English. Alma mater: Fordham University, New York Military Academy, The Kew-Forest School... & Sex: male. Birth: 1961, Kapiolani Medical Center for Women and Children. Religion: Protestantism. Occupation: politician, statesperson. Spouse: Michelle Obama. Languages: Indonesian, English. Alma mater: Columbia University, Harvard Law School, Noelani Elementary School, Occidental College... \\
    \midrule
    Wikiperson\cite{wikiperson} & 45th president of the United States & 44th president of the United States \\
    \midrule
    Weibo\cite{weibo} & 45th President of the United States of America  & Dad, husband, President, citizen.  \\
    \midrule
    Wikidiverse\cite{wikidiverse} & For other uses, see Donald Trump. & Barack, "Obama" redirect here. For other uses, see Barack, Obama, and Barack Obama.  \\
    \midrule
    \textbf{Wiki-S} & Donald John Trump (born June 14, 1946) is an American politician, media personality, and businessman who served as the 45th president of the United States from 2017 to 2021. Trump graduated from the Wharton School.. & Barack Hussein Obama II; born August 4, 1961) is an American politician who served as the 44th president of the United States from 2009 to 2017. A member of the Democratic Party, Obama was the first African-American president of the United States. \\
    \midrule
    \textbf{Wiki-D} & Donald Trump is an American businessman, television personality, and politician who served as the 45th President of the United States from January 2017 to January 2021. Trump was born on June 14, 1946, in Queens, New York City and comes from a wealthy family involved in real estate... & Barack Obama is an American politician and attorney who served as the 44th president of the United States from 2009 to 2017. He is a member of the Democratic Party and was the first African American president of the United States. Before his presidency, Obama served as a senator from Illinois.... \\
    \bottomrule
    \end{tabular}%
  \label{tab:er}%
\end{table}%

    As shown in Table \ref{tab:er}, we analyze the existing entity linking datasets and their methods of entity representation. (It is worth noting that Weibo dataset~\cite{weibo} is primarily designed for Chinese celebrities and does not involve international figures. For the sake of convenient presentation and visual comparison, following the definition of the Weibo dataset, we manually gathered personal profiles of relevant entities from Twitter)
    \begin{itemize}
        \item Wikimel and Richpedia~\citep{baseline_dataset} employ concise attributes from Wikidata. This representation lacks representativeness for entities, as many entities share similar attributes.
        \item Wikiperson~\citep{wikiperson} employs simpler attributes, such as occupation, to represent individuals within it.
        \item Entities in the Weibo~\citep{weibo} dataset are primarily celebrities, and their entity representations are constructed using personal bios written by users on the social platform "Weibo". These bios are written by individual users and represent their personal expressions, which may not accurately reflect the public's perception, leading to semantic inconsistencies.
        \item Wikidiverse~\citep{wikidiverse} utilizes rich multimodal information as entity representation, including both images and text. However, images can deviate from a person's true appearance due to factors like angles and time, lacking real-time accuracy.
    \end{itemize}

    \subsection{Static Enhancement}
        Previous works \cite{baseline_mel, baseline_dataset} tend to concatenate the main properties of an entity into a piece of text and regard the properties as textual entity representation. However, a set of properties is not clear enough to distinguish the linking entity from candidate entities. Different entities may be the same in some fields. As shown in the entity representation in Figure \ref{fig:model}, the properties of \textit{Mike Pompeo} are similar to \textit{Donald Trump} in original representation. Thus it is challenging to distinguish the two people with nearly the same properties.
        Therefore, we leverage Wikipedia to get static enhanced representation, with the textual Wikipedia description from the text on Wikipedia pages.  

        \begin{figure}[pbt]
            \centering
            \includegraphics[width=.6\linewidth]{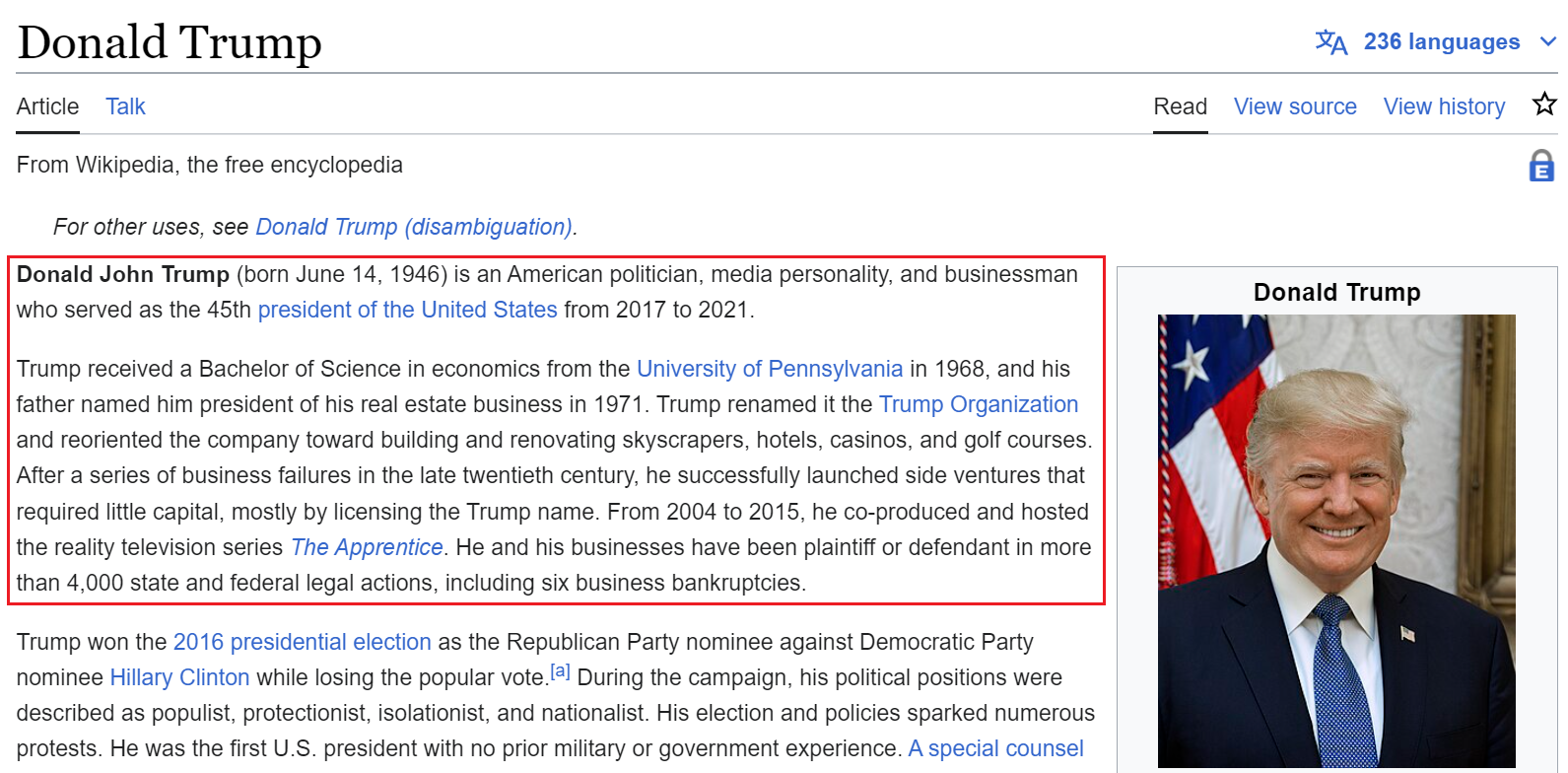}
            \caption{Textual Description of Donald Trump from Wikipedia.}
            \label{fig:wiki}
        \end{figure}

        As shown in Figure \ref{fig:wiki}, the textual description is obtained from the first two paragraphs in a knowledge base, such as Wikipedia\footnote{\url{https://en.wikipedia.org/wiki}} or DBpedia\footnote{\url{https://www.dbpedia.org/}}. We further clean the data by removing special characters, restricting its length, and discarding entities that lack corresponding descriptions in knowledge base. For these entities lacking descriptions, it is unnecessary to link entities that do not have corresponding entries in the knowledge base.
        In detail, we remove 496, 231, and 924 samples from the Richpedia, Wikimel, and Wikidiverse datasets, respectively. The proportion of removed entities to the total number of entities in each dataset was 2.78\%, 1.32\%, and 1.62\%.

    \subsection{Dynamic Enhancement}
        % 静态表示是从维基百科或其他知识库手动收集的，是实体在某个阶段的固定状态和信息。然而，人类对实体的理解随着时间和事件的变化而变化。 例如，乔拜登曾于2009-2017年任美国副总统，2024年则成为了美国总统。在这种情况下，实体的表示也应该随着时间而进行动态调整。表中的传统实体表示方法是僵化且适应性较差，这可能会导致实体身份的偏差和链接错误。 因此，我们进一步提出了一种动态方法来增强实体表示，通过利用飞速发展的大模型对知识的理解能力，以完成对实体表示的动态增强和扩展。
        % Static representations are manually collected from Wikipedia or other knowledge bases and can only represent the entity's state at a specific time. Human understanding of entities changes over time and events. For instance, Donald Trump is no longer the President of the United States in 2023. In such cases, rigid and less adaptable entity representations can lead to errors. 
        % Therefore, we further propose a dynamic method for enhancing entity representations, leveraging the capabilities of ChatGPT to enhance real-time and scalable entity representations based on the evolving understanding of the world.
        Static representations are manually collected from Wikipedia or other knowledge bases and represent the fixed state of the entity at a particular stage. However, human understanding of entities evolves over time and events. For example, Joe Biden served as Vice President of the United States from 2009 to 2017 and became President of the United States in 2024. In such cases, entity representations should also be dynamically adjusted over time. Traditional entity representation methods in Table \ref{tab:er} are rigid and less adaptive, which may lead to biases in entity identity and linking errors. Therefore, we further propose a dynamic method to enhance entity representations by leveraging the rapid development of large models' understanding of knowledge, to accomplish dynamic enhancement and expansion of entity representations.
        ChatGPT~\citep{GPT4} is a powerful large language containing a wealth of knowledge and information about entity cognition. Its knowledge is continuously updated over time and we plan to utilize it to get textual information about entities. Candidate entities are input into ChatGPT for inquiries using the prompt: ``You are a helpful assistant designed to give a comprehensive introduction about people. Who is this one?" The generated response from ChatGPT is shown in Table \ref{tab:er}.

        % 在针对Wikimel数据集中收集到的17391个实体表示中，2142个实体需要进一步的信息进行确实，“It is possible that Edward J. Livernash is a private individual without any notable achievements or public presence.”
        Taking the Wikimel~\citep{baseline_dataset} as an example, among the 17391 entities collected from the dataset, 2142 entities require additional information for verification. For instance, ``Enno Hagenah is a private individual without significant public recognition or fame. Without more specific information, it is difficult to provide a comprehensive introduction."
        % 对于这2142个实体，我们使用多轮对话的方式增加询问的粒度和前置信息，以获得相应的实体信息。
        For these 2142 entities, we employ a multi-turn dialogue approach to enhance granularity in inquiries and provide additional context to acquire relevant entity information.
        % 维基百科中存在关于实体的知识，但是这种知识是静态的，因此我们利用维基百科中的详细知识作为ChatGPT的提示
        The knowledge available about entities in Wikipedia is static in nature. To overcome this limitation, we leverage the detailed information present in Wikipedia as prompts for ChatGPT to get dynamic entity representation.
        % 因此，为了完整性，我们采用从Wikipedia来补齐这部分缺失的实体表示，如“Edward James Livernash, subsequently Edward James de Nivernais (February 14, 1866-2013 June 1, 1938), was an American newspaperman and lawyer who served one term as a U.S. Representative representing the fourth congressional district of California from 1903 to 1905.”
        For example, ``Enno Hagenah is a German politician for the Alliance 90/The Greens. He was elected to the Lower Saxon Landtag in 1998, and was re-elected on two occasions in 2003 and 2008, leaving the Landtag in 2013. Please provide more detailed information."
        In summary, based on the response of ChatGPT, we construct entity representations for 17391, 17804, 57007 entities from Wikimel, Richpedia, Wikidiverse, respectively(i.e. Rich-D, Wiki-D, Diverse-D). 
        The newly built entity representations better reflect the general public's understanding of entities, align closely with their inherent semantics, and facilitate a unified approach to multimodal information and knowledge base.

\section{Methodology}
    \begin{figure*}[h]
        \centering
        \includegraphics[width=\textwidth]{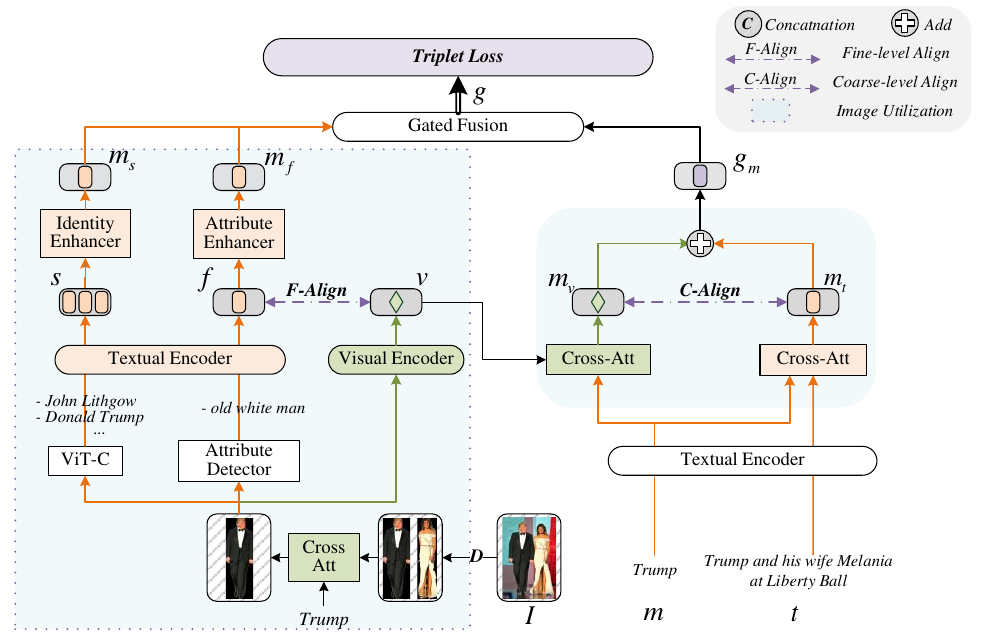}
        \caption{
            Overview of our method.
            Input consists of image $I$, text $t$, and mention $m$. Object detection is applied to extract object feature $d_i$ from image. Facial feature $f$ and identity feature $s$ are retrieved from image.
            % \scriptsize{Textual Encoder}
            % $Q_L, K_L, V_L$ is the linear layer for query, key and value.
        }
        \label{fig:model}
    \end{figure*}

    The overview of our model is shown in Figure \ref{fig:model}. Our method consists of four parts: 
    \textit{\textbf{(1)} Features extraction}. We extract visual features by CLIP image encoder and textual features by CLIP text encoder. We design a non-linear function to map the feature of mention $m$, text $t$, image $v$, and entity representation $E$ into the same feature space. 
    \textit{\textbf{(2)} Image Refinement}. Object detection is applied to extract $l$ objects $\hat{d_1},...\hat{d_{l}}$ from raw image $I$. Furthermore, we get the detected visual object feature $D=d_1,...d_l$ by the image encoder. However, only a few visual objects are related to mention, and thus we apply mention to guide feature extraction of the related visual object $R$. 
    \textit{\textbf{(3)} Visual attribute extraction}. we extract the visual attributes (such as facial feature and identity) and concatenate them into the prompt sentence. Then we leverage prompt sentences as auxiliary information. We obtain the linking entities by measuring cosine similarity between joint multimodal feature $g$ and candidate entities $C_e$ with textual Wikipedia description.
    \textit{\textbf{(4)} Hierarchical contrastive training}. we consider the features of text $t$ and image $v$ as the coarse-level feature because the content is not all related to mention. On the contrary, the feature of mention $m$ and related visual object $d$ are regarded as fine-level feature. To this end, we design coarse-level contrastive loss $L_{c}$ to guide the learning between coarse-grained features (i.e. text $t$ and image $v$). Meanwhile, fine-level contrastive loss $L_{f}$ is designed for fine-grained feature (i,e. mention $m$ and related visual object $d$). 
    
    \subsection{Notation and Task Definition}
        Let $X = \{x_i\}_{i=1}^N $ be a set of $N$ input multimodal samples, with corresponding entities set $Y = \{e_i\}_{i=1}^M$ in a KG. Each input is composed of three parts: $x = \{x_m; x_t; x_v\}$, where $x_t$ denotes a text consisting of a sequence of tokens with length $l_t$, $x_v$ is an image associated with the text, and $x_m$ is a set of mentions in text $x_t$.
        
        A KG is composed of a set of entities $Y = \{e_i\}_{i=1}^M$. Each entity $e_i$ is described by a sequence of entity representations $e_i= u_1, u_2... u_{w}$ (i.e., textual Wikipedia description about the entity). $w$ is the token length of entity representation.
        The MEL task aims to link the ambiguous mention (i.e. query) to the entity in KG by calculating the similarity between the joint mention feature and textual entity representation. The entity with the highest similarity is regarded as the corresponding linking entity. In this process, textual and visual information is used to disambiguate and further clarify semantics. Finally, the KG entity $y$ linked to mention is selected from $\lambda$ candidate entities $C_e= \{e_i\}_{i=1}^\lambda$. 
        It can be formulated as follows:
    	\begin{equation}
    		y = \mathop{argmax} \limits_{\forall e \in C_e} \Gamma(\Phi(x), \Psi(e))
    	\end{equation}
        where $\Phi$ represents the multimodal feature function, $\Psi$ represents the KG entity representation feature function, and $\Gamma()$ produces a similarity score between the multimodal representation and entity representation in KG. $y$ is the predicted linking entity.

        In evaluation, we take cosine similarity to measure the correlation between the joint multimodal feature $g$ and the candidate entities $C_e$. The entity with the highest similarity to $g$ is regarded as the corresponding linking entity. For the example shown in Figure \ref{fig:intro1}, The similarities between the mention and entities are 0.47, 0.81, and 0.13. Therefore, candidate entity \textit{Donald Trump} is chosen as the linked entity.

        Given a sentence $x_t$, mention $x_m$ and Wikipedia description of entity  $x_e$, we follow CLIP \cite{clip_model} to tokenize it into a sequence of word embeddings. Then the special tokens \textit{startoftext} and \textit{endoftext} are added at the beginning and end positions of word embeddings. As a result, with $N$ sentences and $N_e$ candidate entities, we feed sentence representation  $t \in \mathbb{R}^{N \times d}$,  mention representation $m \in \mathbb{R}^{N \times d}$ and enhanced entity representation $g_e \in \mathbb{R}^{N_e \times d}$ into model.

    \subsection{Object-oriented Image}
        Multimodal information in the MEL task generally consists of textual and visual information. Text is artificial, and thus its semantics is intensive. In contrast, visual information tends to be a photo of objects or scenes. There is inevitably some redundant information in the raw image. For an image with multiple mentions, only some interested objects in the image are related to one mention, while the rest are related to the other mention. Some objects are still irrelevant and disturbing, even for an image with a mention. Therefore, refining the raw image by detecting and extracting the related visual object is necessary.
        As shown in Figure \ref{fig:model}, the text is \textit{Trump and his wife Melania at the Liberty Ball}, and the mention is \textit{Trump}. Only the local region of the left person contributes to linking mention \textit{Trump} and entity \textit{Donald Trump} in Wikidata, while the rest of the image is of less significance.
        % However, the rest of the image is an essential clue for other EL, such as linking \textit{Melania (the right person)}.

        % 因此我们将图像按照object oriented的方式进行提取。将完整的图像依据Object Detection结果进行划分
        Therefore, we extract images in an object-oriented manner by dividing the entire image based on object detection results.
        An image $I$ can be decomposed into $l$ objects in local regions. Each mention only describes a part of the image. Therefore, we take an object detector \textbf{\textit{D}}\cite{pixellib} to identify all possible objects in the image. The object visual feature $d$ is extracted through the image encoder of CLIP.
            \begin{align}
                x_d & = \{x_d^i \}_{i=1}^{l} = D(I) \\
                d & = clip(x_d)
            \end{align}
        where $d = \{d_{i}\}_{i=1}^{l}$ and $l$ is the number of the detected objects in the image. 

        Only some objects in the image are related to the mention. Thus, we utilize the visual enhancer to determine related visual feature and obtain vision-enhanced mention feature $m_v$.
            \begin{equation}
                m_d = \text{Cross-Att}(m, d)
                \label{eq:att}
            \end{equation}

    \subsection{Multimodal Enhanced Query}
        We build three enhanced units based on cross-attention to capture dense interaction between mention and multimodal information.

        % \paragraph{Cross-modal enhancer.} 
        %     Given two types of  features, $X$ and $Y$, the  enhanced features  is calculated as:
        %         \begin{equation}
        %             E_n = E_n(X, Y)
        %         \end{equation}
        %     Specifically, the calculation of multi-head attention for hidden state $h_t$ is  first performed on the hybrid key and value:
        %         \begin{equation}
        %             h_t = Softmax(\frac{(W^QX)(W^KY)^T}{\sqrt{d}})(W^VY) 
        %         \end{equation}
        %     Then we utilize the transformer decoder for decode  hidden state $h_t$ with the learnable textual query $Q_v \in \mathbb{R}^{N_v\times d}$:
        %         \begin{equation}
        %             E_n = Decoder(h_t, Q_v)
        %         \end{equation}
        %     In detail, we formulate the attention of transformer decoders in Decoder as:
        %         \begin{equation}
        %             E_n = Softmax(\frac{(W^QQ_v)(W^Kh_t)^T}{\sqrt{d}})(W^Vh_t) 
        %         \end{equation}
        %      where $W^Q \in \mathbb{R}^{d\times d_q}, W^K \in \mathbb{R}^{d\times d_k}, W^V \in \mathbb{R}^{d\times d_v}$ are randomly initialized projection matrices. We set $d_q=d_k=d_v=d/h$. $h$ is the number of heads of attention layer. 

        \subsubsection{Face-enhanced unit}  
        % \footnote{https://github.com/serengil/deepface}
            Deepface\cite{deepface} is applied to explicitly extract facial appearance $x_f$ such as gender, race, age and etc, which are relevant to the attributes of facial feature in the image. $x_f$ is integrated into the sentence such as \textit{Trump, gender: male, race: white, age: 50} to get facial feature $f= \{f_{i}\}_{i=1}^{l}$. The facial feature $f_i$ is related to visual object feature of i-th object of image. Cross attention is applied to get face-enhanced feature $m_f$.
            This attention shares parameters with the attention in Equation \ref{eq:att}. 
            This attention represents which specific object region of the image the mention focuses on, thereby attending to the corresponding facial attributes in that area.
            \begin{equation}
                m_f = \text{Cross-Att}(m, f)
            \end{equation}

        \subsubsection{Identity-enhanced unit}   % Identity
            \begin{figure}[h]
                \centering
                \includegraphics[width=.5\linewidth]{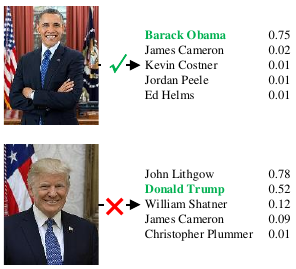}
                \caption{The preprocessing results on ViT, analyzing the similarity with facial images in the library. ViT is pretrained on MS-Celeb1M.}
                \label{fig:celeb}
            \end{figure}
            
            MS-Celeb1M\cite{celeb} is a dataset for recognizing one million celebrities from their face images, where the data is from all the possibly collected face images of this individual on the web.
            % 我们使用在CELEB数据集上进行预训练的ViT对图像进行预处理，分析图像中的人物可能与那些名人的图像相似，为后续entity判断提供辅助信息。
            Images are preprocessed by using ViT \cite{ViT} model pretrained on the face recognition dataset MS-Celeb1M to analyze which celebrities the people in the images may resemble, providing auxiliary information for subsequent entity identification.
            Thus, we could get identity attributes $x_s$ such as \{score: 0.756, label: Barack Obama\} from image. 
            % 如图所示，分析图中人物的相似度列表。但是，由于图像的角度等问题，识别结果不会准确，如图中第二部分，可能会分析错，因此其结果只能作为辅助信息。
            As shown in Figure \ref{fig:celeb}, we analyze the similarity list of people in the image. However, due to issues such as the angle of the image, the identification results may not be accurate, as illustrated in the second part of Figure \ref{fig:celeb}. Therefore, its results can only serve as auxiliary information. To keep accuracy and reduce error propagation, we only use identity information with confidence scores greater than 0.5 and concatenate them into $x_s$.
            We integrate and encode $x_s$ to get identity feature $s$. Then cross attention is applied to get the identity-enhanced feature $m_s$.
                \begin{equation}
                    m_s = \text{Cross-Att}(m, s)
                \end{equation}
            where $s \in \mathbb{R}^{n_s\times d}$ and $m_s \in \mathbb{R}^{1 \times d}$, $d$ indicates the hidden dimension and $n_s$ is the number of identity. This attention shares parameters with the attention in Equation \ref{eq:att}.

        \subsubsection{Text-enhanced unit} 
            The unit is designed to model text-to-mention correlations. It takes two groups of feature sentence representation  $t \in \mathbb{R}^{N \times d}$,  mention representation $m \in \mathbb{R}^{N \times d}$ as inputs to get text-enhanced features:
                \begin{equation}
                    m_t = \text{Cross-Att}(m, t)
                \end{equation}

    \subsection{Gated Feature Fusion}
        We get the feature of mention $m$, text-enhanced mention $m_t$, vision-enhanced mention $m_v$ and attribute-enhanced mention $m_s$ and apply gated fusion to restrain noise in $m_s$. 
    	\begin{align}
    		g_m &= m + m_t + m_v \\
                g &= ((1-\epsilon)g_m + \epsilon (m_f + m_s)) W_g  + b_g
    	\end{align}
        where $W_g \in \mathbb{R}^{2d \times d_{m}}$, $d$ is hidden dimension, $d_{m}$ is the dimension of joint feature $g$. $b_g$ and $\epsilon$ is learnable parameters.

    \subsection{Hierarchical Contrastive Training}
        \subsubsection{Contrastive Loss}
        We apply hierarchical contrastive learning to improve the quality of joint multimodal feature $g$. We design contrastive loss at different granularity: coarse-level contrastive loss and fine-level contrastive loss.
        \textbf{(1)} \textit{Coarse-level} contrastive loss is designed for coarse-grained information, i.e., text and image. Text and image consist of not only information related to mention but also redundant objects or noise.
        Although the alignment between text and image is coarse due to information granularity, it represents alignment of overall semantics, which is still necessary.
        \textbf{(2)} \textit{Fine-level} contrastive loss is designed for the fine-grained information: mention and related visual object. Mention is the subject name that we are concerned about, and the related visual object is the local region described by mention. The noise of mention and related visual objects has been greatly alleviated by the refinement process. Thus this contrastive learning is considered fine-level.
        
        We take Mean-Shifted Contrastive Loss (MSC) \cite{loss} to guide hierarchical contrastive learning on (1)fine-level alignment for facial feature $f$ and object visual feature $d$; (2)coarse-level alignment for text-enhanced feature $m_t$ and vision-enhanced feature $m_v$.
        Specifically, the MSC loss for two information $f_1, f_2$ from an augmented mini-batch of size 2B, is defined as follows. 
        % $f_{con}$ is the concatenation of information $f_1, f_2$, i.e., $f_{con} = [f_1, f_2]$. $\tau$ denotes a temperature hyperparameter.
	\begin{equation}
		\begin{aligned}
			\mathcal{L}_{m} &=  - \log( \\ 
			&\frac{exp(f_1 \cdot f_2 / \tau)}{\sum_{i=1}^{2B} \mathbbm{1}_{i \neq m}  \cdot exp(f_{c} \times f_{c}^T / \tau) - exp(f_1 \cdot f_2 / \tau)})
		\end{aligned}
	\end{equation}
 
        Tthe coarse-level contrastive loss $\mathcal{L}_{c}$ and fine-level contrastive loss $\mathcal{L}_{f}$ are defined as: 
    	\begin{align}
    		\mathcal{L}_{c} & = \mathcal{L}_{m}(m_t, m_v) \\
    		\mathcal{L}_{f} & = \mathcal{L}_{m}(f, d)
    	\end{align}

        \subsubsection{Triplet Loss}  
        During training, we attempt to maximize the similarity between the mention and its corresponding entity (positive sample $\theta_+$) in KG and minimize the similarity between the mention and other entities (negative sample $\theta_-$).
        Thus we take triplet loss\cite{triplet} to maximize the distance between multimodal feature $g$ and negative samples while minimizing the distance between $g$ and positive samples. The triplet loss $\mathcal{L}_t$ is defined as:
    	\begin{equation}
    		\mathcal{L}_t = max( \Gamma(g, \theta_+) - \Gamma(g, \theta_-) + \varepsilon, 0)
    	\end{equation}
        $\Gamma$ is cosine similarity between features. $\varepsilon$ denotes margin, a hyperparameter for learning.

	\subsubsection{Training Loss}
        The overall loss is the combination of triplet loss and contrastive loss. $\alpha$ is weight parameter between two contrastive losses. $\beta$ is the coefficient factor for balancing the losses of two tasks. The results under different $\lambda_c$ and $\lambda_t$ is shown in Figure \ref{fig:details}  
            \begin{equation}
    		\mathcal{L} = \mathcal{L}_f + \alpha \mathcal{L}_c + \beta \mathcal{L}_t
    	\end{equation}

\section{Experiments}
    We first present experimental settings, including datasets, baseline methods, parameter settings, and evaluation metrics. Then, we compare our model DWE+ with state-of-the-art MEL method. Furthermore, we conduct the ablation study and detailed analysis of the components. Finally, we perform a detailed case study and analysis.

    \subsection{Datasets and Settings}
        We take three public datasets: Richpedia\cite{Richpedia}, Wikimel\cite{baseline_dataset}, and Wikidiverse\cite{wikidiverse} as benchmark. Table \ref{tab:statistics} shows the statistics of datasets. 
            \begin{table}[pbt]
            \caption{The statistics of three multimodal datasets}
              \centering
                \begin{tabular}{lcccc}
                \toprule
                Dataset & Sample & Entity & Mention & Text length \\
                \midrule
                Richpedia & 17805 & 17804 & 18752 & 13.6 \\
                Wikimel & 18880 & 17391 & 25846 & 8.2 \\
                Wikidiverse & 13765 & 57007 & 16097 & 10.1 \\
                \bottomrule
                \end{tabular}%
              \label{tab:statistics}
            \end{table}%

	\subsubsection{Metrics}
    	We use T@k (Top-k) accuracy as the metric:
    	\begin{equation}
    		% Acc_{t@k} = \frac{1}{N}\sum_{i=1}^{N}\eta(t_i \in y_i^k)
                Acc_{top\text{-}k} = \frac{1}{N}\sum_{i=1}^{N}\eta \{ I(cos(g, gt), cos(g, C_e)) \leq k \}
    	\end{equation}
            where $N$ represents the total number of samples, and $\eta$ is the indicator function. When the receiving condition is satisfied, $\eta$ is set to 1, and 0 otherwise. $gt$ is ground truth entity while $C_e$ is a set of candidate entities. $cos$ means cosine similarity function. $I$ is function to calculate the rank of similarity between joint feature $g$ and ground truth $gt$ among a set of candidate entities $C_e$.

	\subsubsection{Hyperparameters and Training Details}
\begin{figure}[htbp]
                \centering
                \subfloat[]{\includegraphics[width=.45\linewidth]{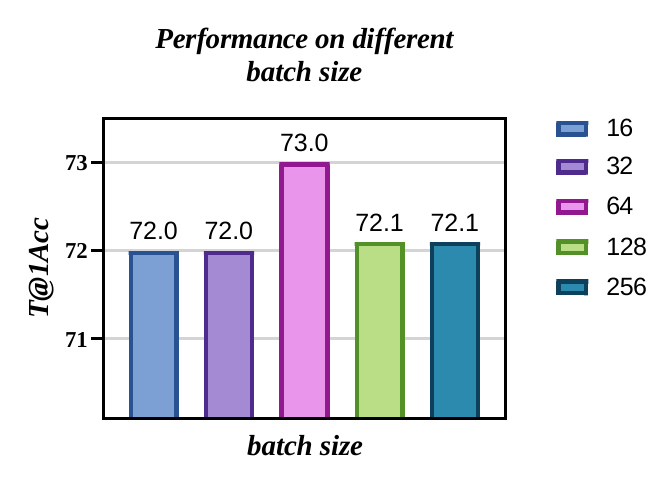}}
                \subfloat[]{\includegraphics[width=.45\linewidth]{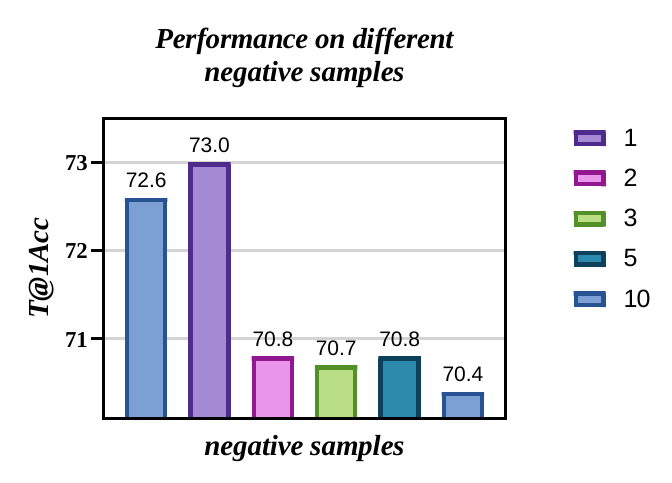}}\\
                \subfloat[]{\includegraphics[width=.45\linewidth]{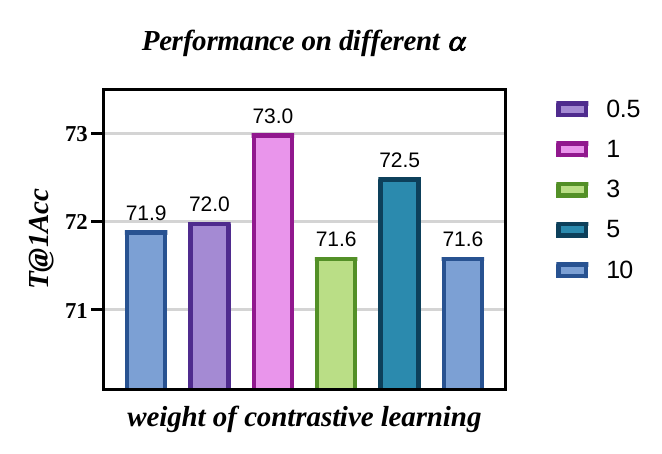}}
                \subfloat[]{\includegraphics[width=.45\linewidth]{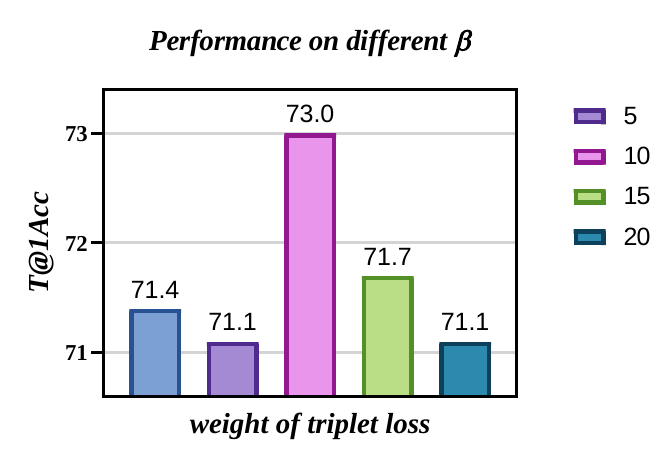}}
                \caption{T@1 accuracy on Wiki-S under different hyperparameters: (a)in batch negative number (b)batch size (c)weight of contrastive loss $\alpha$ (d)weight of triplet loss $\beta$.}
                \label{fig:details}
            \end{figure}

            During the experiments, the dimension of text representation $t$ and visual representation $v$ are set to 512, and the heads of Multi-head attention is set to 8.  The dropout, training epoch, evaluating steps, weight decay, and triplet loss interval are set to 0.4, 300, 2000, 0.001, and 0.5, respectively. We optimize the parameters using AdamW \cite{adam} optimizer with a batch size of 64, the learning rate of $5 \times 10^{-5}$. All the experiments are processed on RTX3070Ti and Pytorch 2.0.
            The T@1 accuracy under different $\alpha,\beta$ is shown in Figure \ref{fig:details}. According to the experimental results, we set parameters $\alpha=1$ and $\beta=10$.

        \subsubsection{Candidate Retrieval}
            Following previous work\cite{baseline_dataset, wikidiverse}, we set the number $\lambda$ of candidate entities to 100. The strategy of candidate entity selection varies on different datasets. In Richpedia and Wikimel, we apply \textit{fuzz}\footnote{https://github.com/seatgeek/fuzzywuzzy} to search 100 candidate entities whose names are similar with mention. In Wikidiverse, the dataset provides 10 similar entities. We first divide entities into different collections according to entity types, such as person, location, and so on. According to mention type, we use \textit{fuzz} to search 90 entities with names similar to mention.

        \subsubsection{Negative Samples}
            We employ two negative sampling strategies for triplet loss: hard negatives and in-batch negatives. The hard negatives are the candidate entities retrieved in the Candidate Retrieval step except for the gold entity. The in-batch negatives are gold entities of other $B\text{-}1$ mentions in the batch, where $B$ is batch size.
            As shown in Figure \ref{fig:details}, the performance of model is associated with the number of in-batch negatives and batch size. Thus we set the number of in-batch negatives to 1 and set batch size to 64.
            % \begin{figure}[h]
            %     \centering
            %     \includegraphics[width=\linewidth]{Figures/details.png}
            %     \caption{T@1 accuracy under different hyperparameters: (1)in batch negative number (2)batch size (3)weight of contrastive loss $\lambda_c$ (4)weight of triplet loss $\lambda_t$.}
            %     \label{fig:details}
            % \end{figure}

    \subsection{Baseline}
        we conduct comparative experiments to evaluate the effectiveness of our model on the original entity representation (combination of properties) for comparison. 
        
        \begin{itemize}
            \item BLINK\cite{BLINK} is a BERT-based entity linking model with a two-stage zero-shot linking.
            \item BERT \cite{BERT} stacks several layers of transformer to encode each token in text.
            \item ARNN \cite{ARNN} utilizes Attention-RNN to predict associations with candidate entity textual features.
            \item DZMNED \cite{DZMNED} takes concatenated multimodal attention mechanism to fuse visual, textual, and character features of mentions.
            \item JMEL \cite{jmel} utilizes fully connected layers to project the visual and textual features into an implicit joint space.
            \item MEL-HI \cite{MELHI} uses multiple attention to get richer information and refine the negative impact of noisy images.
            \item HieCoAtt \cite{HieCoAtt} is a multimodal fusion mechanism, using alternating co-attention and three textual levels (tokens, phrases, and sentences) to calculate co-attention maps.
            \item MMEL \cite{MMEL} is a joint feature extraction module to learn the representations of context and entity candidates, from both the visual and textual perspectives. 
            \item GHMFC \cite{baseline_mel} is a advanced baseline proposed by Wang et.al \cite{baseline_mel}. It takes the gated multimodal fusion and novel attention mechanism to link multimodal entities.
            % \item GHMFC$^\dagger$ \cite{baseline_mel} is the reproduction of the GHMFC \cite{baseline_mel} with textual Wikipedia description as enhanced entity representation.
            \item CLIP-text \cite{clip_model} only uses textual information such as text and mention in the dataset. It mainly focuses on the ability to understand textual relationships among text, mention, and entity representation.
            \item CLIP \cite{clip_model} take both textual and visual features into consideration. The model concatenates multimodal features and calculates the similarity between features and ground truth.
            
        \end{itemize}

\begin{table*}[htbp]
    \caption{MEL results of the compared models on Wikimel.
    }
    \centering
    \begin{tabular}{l|llll}
    \toprule
    \multicolumn{1}{c|}{\multirow{2}[2]{*}{Models}}  & \multicolumn{4}{c}{\textbf{Wiki}}  \\
          & T@1 & T@5 & T@10 & T@20 \\
    \midrule
    BLINK & 30.8  & 44.6  & 56.7  & 66.4  \\
    DZMNED & 30.9  & 50.7  & 56.9  & 65.1  \\
    JMEL  & 31.3  & 49.4  & 57.9  & 64.8  \\
    BERT  & 31.7  & 48.8  & 57.8  & 70.3  \\
    ARNN  & 32.0  & 45.8  & 56.6  & 65.0  \\
    MEL-HI & 38.7  & 55.1  & 65.2  & 75.7  \\
    HieCoAtt & 40.5  & 57.6  & 69.6  & 78.6  \\
    GHMFC & 43.6  & 64.0  & 74.4  & 85.8  \\
    CLIP  & 36.1  & 81.3  & 92.8  & 98.3  \\
    DWE  & 44.5  & 66.3  & 80.7  & 93.1  \\
    \midrule
    DWE+ & 44.9  & 67.2  & 81.8  & 93.8  \\
    \bottomrule
    \end{tabular}%
    \label{tab:experiment2}
\end{table*}%

% Table generated by Excel2LaTeX from sheet 'Sheet2'
\begin{table*}[htbp]
    \caption{MEL results of the compared models on Richpedia.}
  \centering
    \begin{tabular}{l|llll}
    \toprule
    \multicolumn{1}{c|}{\multirow{2}[2]{*}{Models}} & \multicolumn{4}{c}{\textbf{Rich}}   \\
          & T@1 & T@5 & T@10 & T@20 \\
    \midrule
    BLINK & 30.8  & 38.8  & 44.5  & 53.6  \\
    DZMNED & 29.5  & 41.6  & 45.8  & 55.2  \\
    JMEL  & 29.6  & 42.3  & 46.6  & 54.1  \\
    BERT  & 31.6  & 42.0  & 47.6  & 57.3  \\
    ARNN  & 31.2  & 39.3  & 45.9  & 54.5  \\
    MEL-HI & 34.9  & 43.1  & 50.6  & 58.4  \\
    HieCoAtt & 37.2  & 46.8  & 54.2  & 62.4  \\
    GHMFC & 38.7  & 50.9  & 58.5  & 66.7  \\
    CLIP  & 60.4  & 96.1  & 98.3  & 99.2  \\
    DWE   & 64.6  & 96.6  & 98.0  & 99.0  \\
    \midrule
    DWE+  & 67.0  & 97.1  & 98.6  & 99.5  \\
    \bottomrule
    \end{tabular}%
    \label{tab:experiment1}
\end{table*}%

% Table generated by Excel2LaTeX from sheet 'Sheet2'
\begin{table*}[htbp]
    \caption{MEL results of the compared models on Wikidiverse.
    }
    \centering
    \begin{tabular}{l|llll}
    \toprule
    \multicolumn{1}{c|}{\multirow{2}[2]{*}{Models}}   & \multicolumn{4}{c}{\textbf{Diverse}} \\
          & T@1 & T@5 & T@10 & T@20 \\
    \midrule
    JMEL  & 21.9  & 54.5  & 69.9  & 76.3  \\
    BERT  & 22.2  & 53.8  & 69.8  & 82.8  \\
    ARNN  & 22.4  & 50.5  & 68.4  & 76.6  \\
    MEL-HI & 27.1  & 60.7  & 78.7  & 89.2  \\
    HieCoAtt & 28.4  & 63.5  & 84.0  & 92.6  \\
    GHMFC & 46.0     & 77.5   & -     & - \\
    CLIP  & 42.4  & 80.5  & 91.7  & 96.6  \\
    DWE   & 46.6  & 80.9  & 91.8  & 96.7  \\
    \midrule
    DWE+  & 47.1  & 81.3  & 92.0  & 96.9  \\
    \bottomrule
    \end{tabular}%
    \label{tab:experiment3}
\end{table*}%

    \subsection{Results on original datasets}
        On Richpedia\cite{Richpedia}, Wikimel\cite{baseline_dataset} and Wikidiverse\cite{wikidiverse} datasets, we compare the proposed model with several competing approaches and show the MEL results of the compared models in terms of T@1, 5, 10, 20 accuracy (\%) from $\lambda$ candidate entities in Table \ref{tab:experiment1},\ref{tab:experiment2} and \ref{tab:experiment3}. Following the setting in \cite{baseline_mel}, $\lambda$ is set to 100.
        We not only report the best performance but also the standard deviation (STDEV) over 10 runs with random initialization. The best results are highlighted in bold. Rich, Wiki, and Diverse are Richpedia, Wikimel, and Wikidiverse dataset, respectively.
        
        Under previous entity representation, our method could reach 64.6\%, 44.7\%, and 47.5\% on three datasets.
        % Besides, under enhanced entity representation, our method also achieves state-of-the-art (SOTA) performance by obtaining T@1 accuracy of 72.5\%, 72.8\%, and 51.2\%, respectively, which demonstrates the superiority over the previous state-of-the-art method.        
        Apparently, with pre-trained on massive multimodal data, the text and image encoder of clip facilitates the understanding of complicated multimodal information. For the gated fusion, it benefits models by restraining noise while using information. Besides, the enhancer for text and vision contributes to fusion between features.

        % \textbf{Enhanced Entity Representation}. Previous works \cite{baseline_dataset,baseline_mel} consider the properties as entity representation, which is ambiguous and limited. we take textual Wikipedia description as enhanced representation. Results show that our method is not only useful for previous entity representation but also the enhanced representation. Besides, on CLIP and our method, the improvement over different representations shows that the enhanced representation is helpful for disambiguation. In detail, on Richpedia, by using enhanced representation, CLIP obtain the 3.5\% improvement and our method obtains 4.9\% improvement. The boost proves that the enhanced entity representation makes the entity more distinctive.

    \subsection{Result on enhanced dataset}
    Original entity representation is ambiguous and is not consistent with entity semantics. Thus, we propose the enhanced entity representation and conduct extensive experiments on it. 
    % 我们将我们的DWE+方法分别在两类数据集上进行了extensive experiment：静态增强的Rich-S、Wiki-S和Diverse-S以及动态增强的Rich-D、Wiki-D和Diverse-D
    % 如表sta和dyn所示，我们的方法在两类增强数据集上均优于基线方法，展现了我们方法的有效性。在Static数据集上，我们的方法分别在Rich-S、Wiki-S、Diverse-S上达到了72.5， 72.8和51.2的T@1准确率。在Dynamic数据集上，我们的方法分别在Rich-D，Wiki-D和Diverse-D上达到了65.1、68.1和53.4的T@1准确率。
    We conduct extensive experiments on two types of datasets: statically enhanced datasets Rich-S, Wiki-S, and Diverse-S, and dynamically enhanced datasets Rich-D, Wiki-D, and Diverse-D. As shown in Tables \ref{tab:static} and Table \ref{tab:dynamic} our DWE+ method outperforms baseline methods on both types of enhanced datasets, demonstrating its effectiveness. On static datasets, our method achieves T@1 accuracy of 67.0\%, 72.8\%, and 51.2\% on Rich-S, Wiki-S, and Diverse-S, respectively. On dynamic datasets, our method achieves T@1 accuracy of 65.1\%, 68.1\%, and 53.4\% on Rich-D, Wiki-D, and Diverse-D respectively.
    \subsubsection{Comparison between original dataset and static dataset}
        % 我们的方法在Static增强数据集Rich和Diverse上的性能有稳定的提升，说明我们所提出的Static增强方式更贴合知识库中实体的真实语义，所以在同等情况下，可以更符合人类行文习惯，与多模态输入保持语义的一致。另一方面可以看到在Wiki数据集上我们的方法在Static数据集上的提升较大，根据我们对Wiki数据集的调查，Wiki数据集中存在Entity representation表示严重模糊不清的情况，因此即使在模型能够理解多模态语义的情况下，仍然很难做到多模态信息和实体表示的双端匹配。这部分大的性能差距更进一步说明了我们所提出的固态增强方法的必要性。
        Our method improves performance on the statically enhanced datasets Rich-S and Diverse-S, indicating that our proposed static entity representation aligns better with the true semantics of entities in the knowledge base. Therefore, under equivalent conditions, our method can better adhere to human habits and maintain semantic consistency with multimodal inputs. On the other hand, we observe our method shows a significant improvement in performance between Wiki and Wiki-S. According to our investigation of the Wiki dataset, we find cases where entity representations are severely ambiguous. Therefore, even when the model can understand multimodal semantics, achieving bidirectional matching between multimodal information and entity representations remains challenging. This significant performance gap further shows the necessity of our proposed static enhancement method.
    \subsubsection{Comparison between original dataset and dynamic dataset}
        % 同样地，动态增强的方法也为我们的模型带来了相应的性能提升，提高了准确率，在Wiki-D和Wiki上较大的性能差距同样说明了原有的Wiki数据集中的Entity Representation存在较大的缺陷。
        Similarly, the dynamic enhancement method also brings corresponding performance improvements to our model. The significant performance gap observed on Wiki-D and Wiki demonstrates the flaws in the original Wiki dataset's entity representation.
    
    \subsubsection{Comparison between static dataset and dynamic dataset}
        % 相比于静态数据集，动态数据集的性能均有不同程度的下降，这表明动态增强的 Entity Representation 获得了对实体更加精细的语义捕捉。这也为模型带来了更高的要求，要求模型能够从多模态输入中捕捉到更多关于实体身份等信息。
        Compared to the static datasets, the performance of dynamic datasets shows varying degrees of decline. This indicates that the dynamically enhanced entity representation could capture finer semantics about entities. It also imposes higher demands on the model, requiring it to understand more information about entity identities and other aspects from multimodal inputs.

\begin{table}[tb]
\renewcommand\arraystretch{1.1}
\caption{Experimental results on the dataset with static enhancement.}
  \footnotesize
  \centering
    \begin{tabular}{l|cccc|cccc|cccc}
    \toprule
    \multicolumn{1}{c|}{\multirow{2}[2]{*}{Models}} & \multicolumn{4}{c|}{\textbf{Wiki-S}} & \multicolumn{4}{c|}{\textbf{Rich-S}} & \multicolumn{4}{c}{\textbf{Diverse-S}} \\
          & T@1 & T@5 & T@10 & T@20 & T@1 & T@5 & T@10 & T@20 & T@1 & T@5 & T@10 & T@20 \\
    \midrule
    GHMFC & 36.1  & 81.3  & 92.8  & 98.3  & 33.4  & 75.7  & 84.9  & 93.7  & 34.3  & 70.8  & 82.8  & 92.0  \\
    CLIP-text & 50.2  & 87.4  & 93.5  & 97.2  & 59.2  & 91.1  & 94.6  & 97.4  & 35.8  & 79.6  & 89.1  & 95.3  \\
    CLIP & 58.0  & 95.9  & 98.3  & 99.5  & 63.9  & 96.2  & 98.3  & 99.2  & 37.0  & 80.2  & 90.7  & 96.2  \\
    DWE   & 71.1  & 97.1  & 98.8  & 99.6  & 66.1  & 96.8  & 98.7  & 99.6  & 51.2  & 91.0  & 96.3  & 98.9  \\
    DWE+  & \textbf{73.0 } & \textbf{97.4 } & \textbf{98.9 } & \textbf{99.7 } & \textbf{67.6 } & \textbf{97.3 } & \textbf{98.8 } & \textbf{99.6 } & \textbf{52.5 } & \textbf{91.3 } & \textbf{96.4 } & \textbf{99.0 } \\
    \bottomrule
    \end{tabular}%
  \label{tab:static}%
\end{table}%

        % Table generated by Excel2LaTeX from sheet 'Sheet2'
\begin{table}[tb]
\caption{Experimental results on the dataset with dynamic enhancement.}
\footnotesize
  \centering
    \begin{tabular}{l|cccc|cccc|cccc}
    \toprule
    \multicolumn{1}{c|}{\multirow{2}[2]{*}{Models}} & \multicolumn{4}{c|}{\textbf{Wiki-D}} & \multicolumn{4}{c|}{\textbf{Rich-D}} & \multicolumn{4}{c}{\textbf{Diverse-D}} \\
          & T@1 & T@5 & T@10 & T@20 & T@1 & T@5 & T@10 & T@20 & T@1 & T@5 & T@10 & T@20 \\
    \midrule
    BERT  & 32.0  & 75.7  & 88.2  & 95.5  & 35.5  & 77.7  & 87.8  & 94.3  & 10.3  & 23.9  & 33.8  & 47.3  \\
    GHMFC & 33.3  & 75.9  & 88.4  & 95.0  & 34.6  & 77.0  & 87.1  & 93.8  & 14.8  & 29.9  & 39.3  & 53.8  \\
    CLIP  & 63.0  & 96.2  & 98.6  & 99.6  & 63.5  & 95.4  & 97.5  & 98.6  & 45.7  & 88.9  & 96.4  & 99.3  \\
    DWE   & 66.9  & 97.7  & 99.2  & 99.7  & 64.3  & 95.9  & 97.7  & 99.1  & 51.3  & 93.7  & 98.8  & 99.5  \\
    DWE+  & \textbf{68.1 } & \textbf{98.4 } & \textbf{99.5 } & \textbf{99.9 } & \textbf{65.1 } & \textbf{96.6 } & \textbf{98.3 } & \textbf{99.5 } & \textbf{53.4 } & \textbf{95.5 } & \textbf{99.2 } & \textbf{99.7 } \\
    \bottomrule
    \end{tabular}%
  \label{tab:dynamic}%
\end{table}%

    \subsection{Comparison with the SOTA Generative Method. }
        To investigate the comprehensiveness of the advantages of our model, we compare our model with the SOTA generative model, GEMEL \cite{GEMEL} on accuracy. Smaller values of the quantity of candidate entity $\lambda$ make the task of selecting link entity from $\lambda$ candidates easier.
        Following the configuration of GEMEL, we set $\lambda$ to 16 and the performance is presented in Table \ref{tab:16}. As we can see, our model can still show superiority.
        % In our work, we adopt the setting from previous work\cite{baseline_mel, MMEL} and set $\lambda$ to 100. 
        % The quantity of candidate entity $\lambda$ is an important factor affecting model performance. Smaller values of $\lambda$ make the task of selecting link entities from these $\lambda$ candidates easier, resulting in higher performance. 
        This is due to our method leveraging cross-modal enhancer and alignment to capture dense interaction over multimodal information while GEMEL mainly relies on the ability of LLM. In fact, the LLM of GEMEL cannot truly learn visual information. Instead, it only depends on CLIP to get visual prefix, inevitably resulting in noise and mismatch between text and image. 

        % Compared with previous best approaches, i.e., GHMFC and CLIP, our method also has great advantages. For example, our method obtains  4.9\% on Richpedia, 10.1\% on Wikimel and 12.9\% on Wikidiverse, respectively.

        \begin{table}[tb]
        \caption{T@1 accuracy for the model from 16 candidates.}
            \centering
            \begin{tabular}{lcc}
            \toprule
                  & Wiki-S & Diverse-S \\
            \midrule
            GEMEL    & 75.5  & 82.7  \\
            DWE+     & 81.0  & 87.1  \\
            \bottomrule
            \end{tabular}%
          \label{tab:16}%
        \end{table}%

    \subsection{Ablation Study}
        % 根据我们之前的分析，原始entity representation不足以代表知识库中的实体，存在表示方法上的缺陷，因此我们主要出于便于与传统方法比较的目的在原始entity representation上进行我们方法的验证。更进一步的，对于我们所提出的dynamic enhance dataset：rich-d, wiki-d and diverse-d，由于该方法是动态进行获得,会随着时间和大模型的版本所进行更改，所以为了固定测试，我们对static enhance dataset：rich-s, wiki-s, diverse-s上进行了细致的Ablation测试。
        Our ablation experiments were primarily conducted on the static dataset. Based on our previous analysis, the original entity representation is insufficient to represent entities in the knowledge base, indicating flaws in the representation method. Therefore, for the sake of comparison with traditional methods, we validate our approach on the original entity representation. Furthermore, for our proposed dynamically enhanced datasets: rich-d, wiki-d, and diverse-d, since this method is dynamically acquired and may change over time and with versions of large models. Thus, for experimental consistency, we conduct detailed ablation tests on the static enhanced datasets: rich-s, wiki-s, and diverse-s.
        Next, we will present the ablation experiments of each module and the corresponding analysis.

\begin{table}[ptb]
    \centering
    \caption{Ablation study on Wiki-S ($t$: text; $v$: image; $s$: identity; $f$: facial characteristics). Method 0 represents our model. Method 4 means only keeping mention feature.}
    \begin{tabular}{rrrrrrrrr}
    \toprule
          & \multicolumn{1}{l}{$t$} & \multicolumn{1}{l}{$v$} & \multicolumn{1}{l}{$s$ } & \multicolumn{1}{l}{$f$} & \multicolumn{1}{l}{T@1} & \multicolumn{1}{l}{T@5} & \multicolumn{1}{l}{T@10} & \multicolumn{1}{l}{T@20} \\
    \midrule
    0     & \checkmark     & \checkmark     & \checkmark     & \checkmark     & 72.96  & 97.40  & 98.91  & 99.61  \\
    1     & \checkmark    & \checkmark    & \checkmark    & $\times$     & 72.69  & 97.79  & 98.99  & 99.57  \\
    2     & \checkmark    & \checkmark    & $\times$     & $\times$     & 72.11  & 97.63  & 98.88  & 99.61  \\
    3     & \checkmark    & $\times$     & $\times$     & $\times$     & 68.39  & 95.15  & 97.52  & 98.95  \\
    4     & $\times$     & $\times$     & $\times$     & $\times$     & 60.67  & 88.40  & 92.59  & 96.16  \\
    \bottomrule
    \end{tabular}%
  \label{tab:ablation}%
\end{table}%

% Table generated by Excel2LaTeX from sheet 'Sheet2'
\begin{table}[ptb]
\renewcommand\arraystretch{1.1}
\caption{Ablation study by using refined visual feature $v$ or primitive visual feature which is extracted directly from image over public datasets}
  \centering
    \begin{tabular}{lcccc}
    \toprule
    Features & T@1 & T@5 & T@10 & T@20 \\
    \midrule
    primitive on wiki-s & 64.0  & 96.2  & 98.4  & 99.5  \\
    refined on wiki-s & 73.0  & 97.4  & 98.9  & 99.7 \\
    \midrule\midrule
    primitive on rich-s & 61.2  & 97.2  & 98.8  & 99.6  \\
    refined on rich-s & 67.6  & 97.3  & 98.8  & 99.6 \\
    \bottomrule
    \end{tabular}%
  \label{tab:visual}%
\end{table}%

        \subsubsection{Attribute enhancement}
        The visual attributes extracted from the image depict the entity's outline and some prominent features. The extracted visual attributes from the image include facial features and identity, which play different roles in the model. 
        (1) Facial features: When the entity in the image is a human subject, the extracted facial features serve as a supplement and correction to the feature extractor, such as in cases where the feature extractor fails to recognize a \textit{white male} in the image. As shown in Table \ref{tab:ablation}, by removing the facial features $f$ (method 0$\rightarrow$1), the model's performance decreased by 1.6\%, demonstrating the effectiveness of facial features. 
        (2) Identity: We leverage the ViT \cite{ViT} pretrained on the face recognition dataset MS-Celeb1M to analyze which celebrities the people in the images may resemble.
        Due to accuracy issues, the recognition results of ViT may not be entirely accurate, so this information can only serve as auxiliary information. Additionally, due to its noise, we use a gated fusion to integrate them. By removing (method 1$\rightarrow$2) identity feature $s$, the accuracy decreased by 1.3\%.

        \subsubsection{Refined visual feature}
        The noise in the image brings a challenge to the multimodal model. In this perspective of view, DWE+ takes visual information refinement to suppress the noise in the image. Table \ref{tab:visual} shows the differences between using refined visual feature $v$ and using primitive visual feature that are extracted from the whole image. It could be seen that the refinement brings 2.1\% improvement over T@1 accuracy.
        Besides, Tables \ref{tab:visual} that the effectiveness of the refined visual feature varies on datasets. The 8.8\% improvement on Wikimel(64.0\%$\rightarrow$72.8\%) is more significant than 5.8\% on Richpedia (61.2\%$\rightarrow$67.0\%). The reason is that the image in Wikimel tends to be noisy with multiple objects, while the object number in Richpedia is less, suggesting that the mechanism is more applicable for noisy datasets (such as Wikimel). Apart from this, Table \ref{tab:ablation} shows our method without visual feature $v$: the 3.72\% drop(method 2$\rightarrow$3) of T@1 accuracy shows the importance of vision. 

        % \subsubsection{Contrastive Loss}
        % The 2.3\% decrease of improvement (method2$\rightarrow$3) shows the contrastive loss helps to reduce the degree of difference between two features (e.g. facial feature $f$ and related visual feature $d$, text feature $t$ and image feature $v$). 

        \subsubsection{Hierarchical Contrastive Loss}
        As shown in Table \ref{tab:loss}, DWE+ takes both coarse-level and fine-level contrastive learning, which shows the effectiveness of hierarchical contrastive learning. Afterward, method 6 removes coarse-level contrastive learning, resulting in a 0.69\% drop in T@1 accuracy. In comparison, the replacement of fine-level contrastive learning (i.e., method 7) makes accuracy decline by 0.66\%. It could be seen that coarse-level and fine-level contrastive learning both play a significant role in MEL. In particular, the fine-level information is more important.

        \begin{table}[H]
        \renewcommand\arraystretch{1.1}
		\caption{Results of contrastive learning ablation experiments on Wiki-S ($\mathcal{L}_c$: coarse-level contrastive loss, $\mathcal{L}_f$: fine-level contrastive loss)}
		\label{tab:loss}
		\begin{tabular}{ccccccc}
		\toprule
			   & $\mathcal{L}_f$ & $\mathcal{L}_c$  & T@1 & T@5 & T@10 & T@20 \\ 
		\midrule
			5 & \checkmark   & \checkmark    & 72.96  & 97.40  & 98.91  & 99.61  \\ 
			6 & $\times$     & \checkmark    & 72.27  & 97.36  & 98.80  & 99.73  \\
			7 & \checkmark   & $\times$      & 72.30  & 97.67  & 98.99  & 99.77  \\
		\bottomrule
		\end{tabular}
	\end{table}

    \subsection{Discussion}

        Compared with method 4, which only uses text, the T@1 accuracy of DWE+(method 0) is improved from 58.0\% to 67.0\%, and other metrics are also improved, which indicates the visual information is critical for the MEL task. Based on method 4, method 3 is built by adding image. The T@1 accuracy of method 4 is improved from 58.0\% to 61.8\%, which illustrates that visual information plays a crucial role in disambiguation and is a supplement to textual information 
              
        It is noticed that method 3 removes visual information refinement resulting in T@1 accuracy obviously decreasing by 4.2\%. In our work, visual information refinement first detects all the objects from the raw image, even if the object may not be what we are concerned about. Afterward, the attention mechanism and contrastive learning help the method focus on the related visual object among all the detected objects. This process effectively refines most of the visual noise, thereby facilitating the alignment of textual and visual information.
        In summary, according to Table \ref{tab:ablation}, the key to performance improvement in DWE+ model lies in two aspects: (1) Visual information refinement helps to filter the visual noise (2) the mention encoding helps to determine the subject of task. Based on these components, our model could get 9\% improvement on T@1 accuracy compared to method 5. The ablation experiment shows the effectiveness of the proposed components and mechanism.

\begin{table*}[pbt]
        \renewcommand\arraystretch{1.1}
		\caption{Multimodal entity linking cases. ER is the static enhanced entity representation. Top-5 entities are selected from $\lambda$ candidate entities. We take the bold blue text to show the correct entity. Case 1 and 2 are multiple object samples for different mentions with the same image and text.}
		\label{tab:casestudy}
		\tiny
		\begin{tabular}{p{0.7cm}p{2.2cm}p{2.2cm}p{2.2cm}p{2.2cm}p{2.2cm}}
			\toprule
		Case              & \makebox[2.2cm][c]{1}  & \makebox[2.2cm][c]{2} & \makebox[2.2cm][c]{3} & \makebox[2.2cm][c]{4} & \makebox[2.2cm][c]{5}                   \\
			\midrule
		Image             
			& \makebox[2.2cm][c]{\begin{minipage}[c]{0.18\columnwidth}\centering{\includegraphics[width=\textwidth]{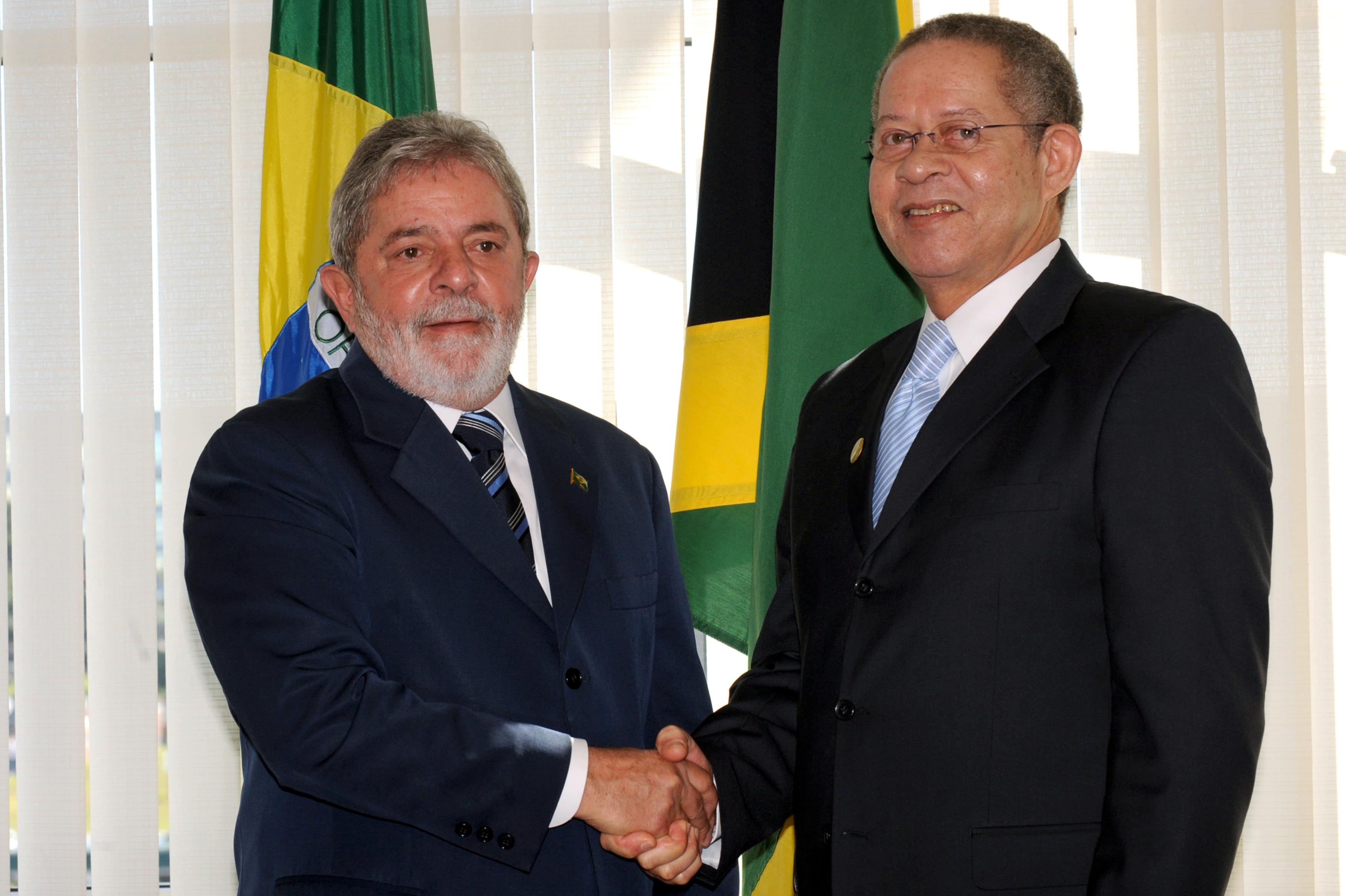}}\end{minipage} }
			& \makebox[2.2cm][c]{\begin{minipage}[c]{0.18\columnwidth}\centering{\includegraphics[width=\textwidth]{utils/42193.jpg}}\end{minipage} }                       
			& \makebox[2.2cm][c]{\begin{minipage}[c]{0.18\columnwidth}\centering{\includegraphics[width=\textwidth]{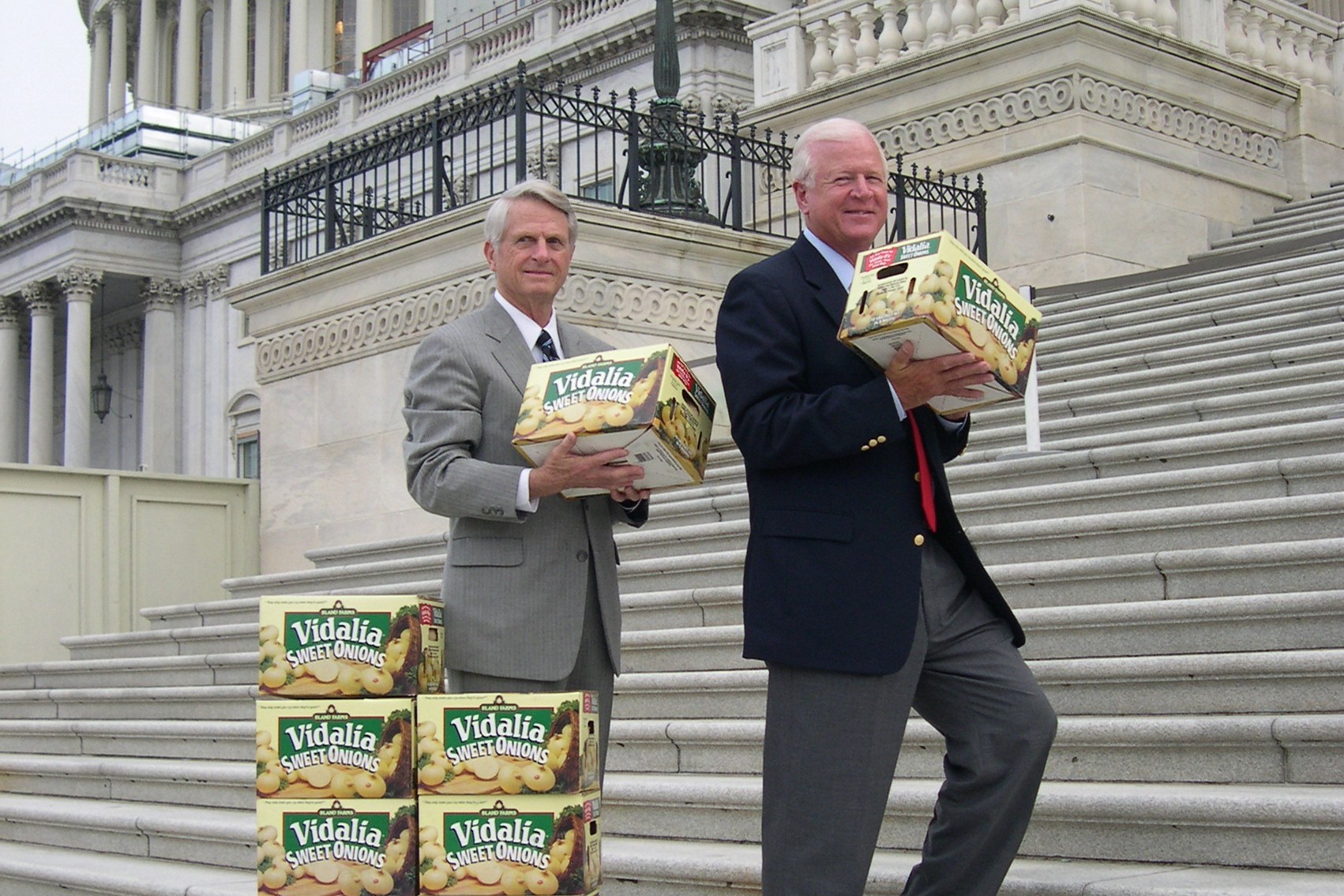}}\end{minipage} }
			& \makebox[2.2cm][c]{\begin{minipage}[c]{0.15\columnwidth}\centering{\includegraphics[height=\textwidth]{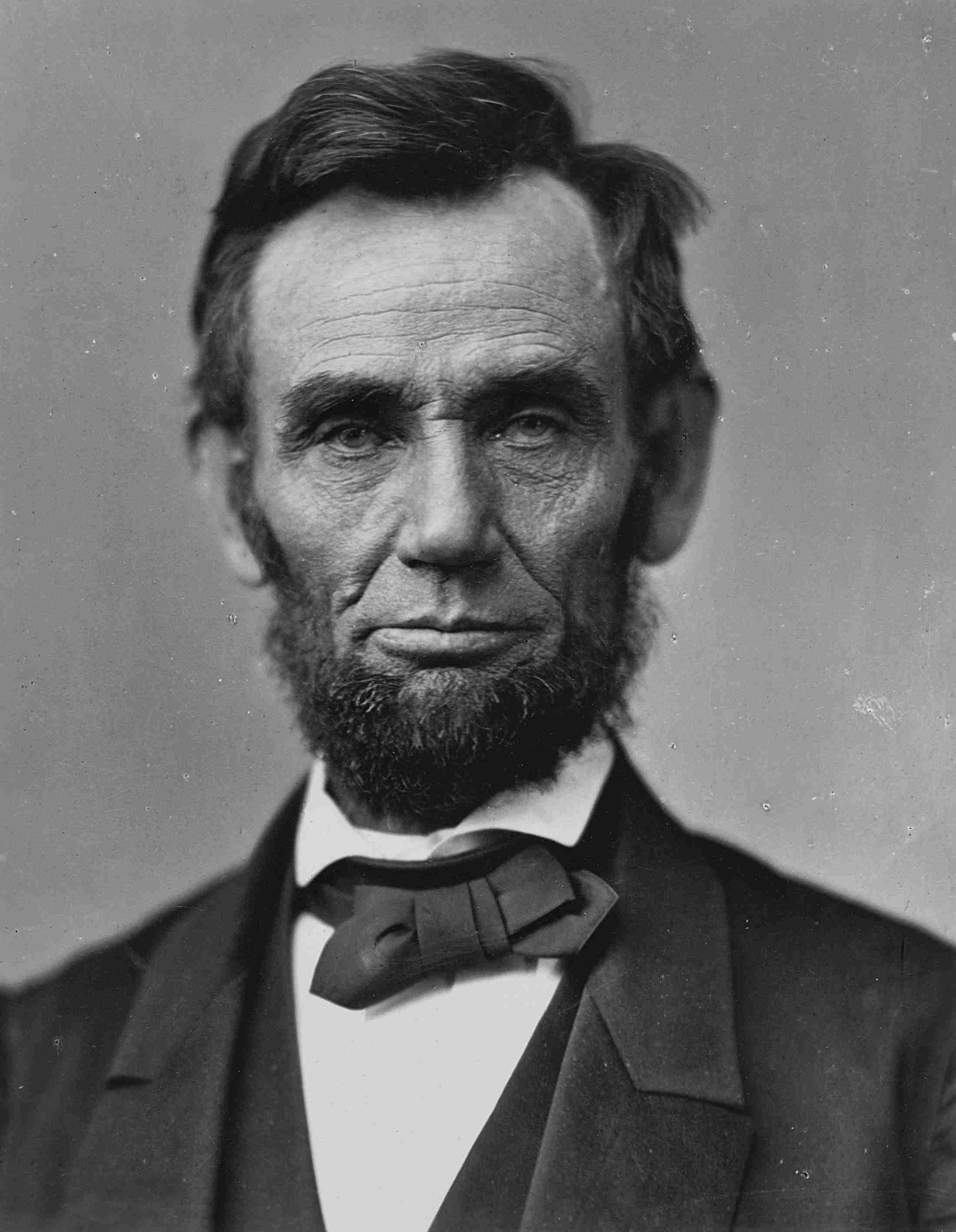}}\end{minipage} }    
			& \makebox[2.2cm][c]{\begin{minipage}[c]{0.15\columnwidth}\centering{\includegraphics[height=\textwidth]{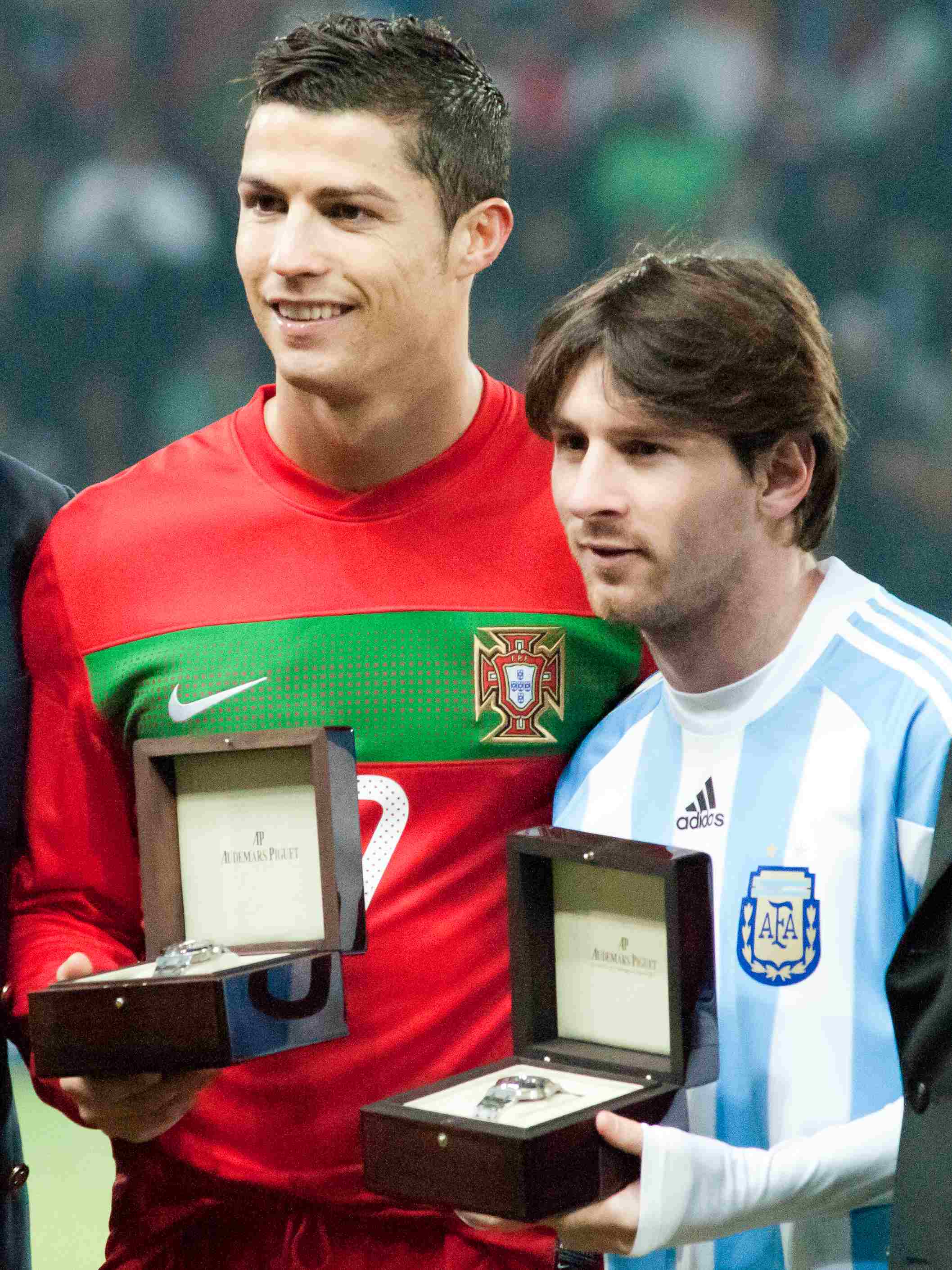}}\end{minipage} }\\
		Text              
			& Golding with the President of Brazil, Lula .         
			& Golding with the President of Brazil, Lula .         
			& Miller and Saxby Chambliss in 2004           
			& Lincoln sought to heal the war-torn nation through reconciliation  
			& Messi has scored over 700 senior career goals for club and country \\
		Mention           & Golding       & Lula              & Miller      & Lincoln         & Messi        \\
		Entity            & Bruce Golding & Luiz Inácio Lula  & Zell Miller & Abraham Lincoln & Lionel Messi \\
		QID 			  & Q315662       & Q37181                    & Q187516     & Q91          	& Q615         \\
		ER  
			& Orette Bruce Golding (born 5 December 1947) is a former Jamaican politician who served as eighth Prime Minister of Jamaica from 11 September 2007 to 23 October 2011..
			& Luiz Inácio Lula also known as Lula  or simply Lula, is a Brazilian politician who is the 39th and current President of Brazil since 2023... 
			& Zell Bryan Miller (February 24, 1932 -2013 March 23, 2018) was an American author and politician from the state of Georgia..
			& Abraham Lincoln (February 12, 1809 -2013 April 15, 1865) was an American lawyer and statesman who served as the 16th president of the United States from 1861 to 1865.. 
			& Lionel Andr\u00e9s Messi (born 24 June 1987), also known as Leo Messi, is an Argentine professional footballer who plays as a forward for Ligue 1 club Paris Saint-Germain. \\
		T@1 Hit   & \textcolor{blue}{\checkmark} &  \textcolor{red}{$\times$}  & \textcolor{blue}{\checkmark} & \textcolor{red}{$\times$} & \textcolor{blue}{\checkmark} \\
        \tabincell{l}{Prediction \\ (Top-5)}
		& \tabincell{l}{\textcolor{blue}{\textbf{Bruce Golding 0.93}}		\\ William Golding 0.27				\\ Jack Golding 0.26	\\ Henry Golding 0.19					\\ Suzanne D. Golding 0.06			}         
		& \tabincell{l}{Fábio Luís Lula  0.71				\\ \textcolor{blue}{\textbf{Luiz Inácio Lula  0.70}}		\\ Marcos Claudio Lula  0.33			\\ Rosângela Lula  0.13					\\ Maria de Lurdes  0.09		}           
		& \tabincell{l}{\textcolor{blue}{\textbf{Sienna Miller 0.79}} 	\\ Marisa Miller 0.61				\\ Kristine Miller 0.56		\\ Ezra Miller 0.31					\\ Dean Miller 0.30		}	       
		& \tabincell{l}{Curt Lincoln 0.53				\\ Lincoln Chafee 0.48				\\ Bill Clinton 0.44		\\ \textcolor{blue}{\textbf{Abraham Lincoln 0.39}} 	\\ Hillary Clinton 0.36	}          
		& \tabincell{l}{\textcolor{blue}{\textbf{Lionel Messi 0.85}} 		\\ Antonello da Messina 0.76		\\ Charles Messier 0.56		\\ Massiv	0.29 					\\ Ramesses II 0.06		}      \\
			\bottomrule
		\end{tabular}
        \label{tab:casestudy}
	\end{table*}

        Besides, the improvement is mainly about the T@1 accuracy from model 5 to model 1. Method 5 has a good feature learning ability because of the pretraining of CLIP. On this basis, the addition of image plays the role of fine-grained information. The text could be seen as a coarse-grained text. It can help to determine the positive sample within a certain set of candidate entities. The fine-grained information (image) contributes to selecting the most likely sample from a certain set.
        
        % % 讲er的作用
        % As shown in Table \ref{tab:experiment}, compared with the multimodal SOTA method GHMFC\cite{baseline_mel}, DWE has a T@1 accuracy increment of 22.4\% on Wikimel and 26.4\% on Richpedia, respectively. GHMFC considers the entity properties as entity representation, which is ambiguous. Thus we reproduce the GHMFC$^\dagger$ on the Wikimel and Richpedia with new entity representation to evaluate the effectiveness. It could be seen that T@5, 10, and 20 accuracy is greatly improved by using static enhanced entity representation. Compared with GHMFC, the performance of GHMFC$^\dagger$ has been greatly improved except for the T@1 accuracy. This result indicates that new entity representation makes the entity more distinctive. However, as for T@1 accuracy, the capacity of GHMFC becomes the bottleneck, i.e., it can not distinguish samples accurately, resulting in a drop in T@1 accuracy. The analysis has also been validated by our experiments in Table \ref{tab:experiment}.

    \subsection{Case Study}
        We further analyze the effectiveness of our model DWE+ by cases in Table \ref{tab:casestudy}, including correct and wrong predictions. 
        Case 1,2,3 are from Wikimel, and case 4,5 are from Richpedia.
        As shown in Table \ref{tab:casestudy}, The text and image in case 1 and 2 are the same, except for the mention. Thus it is challenging to identify a unique entity from images with multiple objects because the interference of other objects is strong. That is the reason that the accuracy of T@5, 10, 20 is much higher than that T@1 in experiments. 

        We first analyze the \textbf{correct} predictions. The correctness of case 1 and 4 shows the case with multiple visual objects on different datasets. As for the correct case 3, the mention is \textit{Miller} with two old men in image. The result shows the effectiveness of the attributes. The model could mine the relationship between \textit{two old men}, and entity representation of \textit{Sienna Miller }, while the others are female.
        On the other hand, \textbf{wrong} predictions can still show the effectiveness of our model. In case 2, the visual and textual information contains multiple objects: \textit{Bruce Golding} and \textit{Luiz In acio Lula da Silva}. It is difficult to distinguish these two mentions by the same multimodal information. However, it could be seen that the similarity of \textit{Fabio Lu is Lula da Silva} and \textit{Luiz In acio Lula da Silva} is much larger than the rest of candidate entity, which still shows learning ability.
        
        The above cases illustrate the ability of DWE+ to mine the semantic relationship between textual information (e.g., text and mention) and entity representation. Besides, the supplementary visual feature and attribute extracted from image are leveraged for further disambiguation.
        In summary, the proposed mechanism helps DWE+ to understand multimodal information and learn joint feature for entity linking.
        
\section{Conclusion}
    
    This paper addresses the challenges in Multimodal Entity Linking (MEL) by proposing a novel Dual-Way Enhanced framework(DWE+). We observe that existing MEL approaches suffer from insufficient utilization of images and inconsistency between entity representation and real-world semantics. To overcome these challenges, we introduce fine-grained image features extraction and enhanced entity representations. Our contributions include the exploration of multimodal entity linking in a neural matching formulation, the introduction of fine-grained image attributes for feature enhancement, and the optimization of entity representations to better signify entity semantics in the knowledge base. We release the optimized datasets to the public, providing valuable resources for future research in the field of entity linking.
    Extensive experiments on three public datasets show the effectiveness of DWE. Furthermore, two optimized entity representation approaches enhance the Richpedia, Wikimel, and Wikidiverse datasets. The enhanced entity representation is more representative and no longer restricts the learning of multimodal information.

\begin{acks}
This work was supported by the Hunan Provincial Natural Science Foundation Project (No.2022JJ30668, 2022JJ30046) and the science and technology innovation program of Hunan province under grant No. 2021GK2001. This work was supported by the National Natural Science Foundation of China (No.72188101, No.62302144).
\end{acks}

\bibliographystyle{ACM-Reference-Format}
\bibliography{main}

\end{document}